\documentclass[10pt,twocolumn,letterpaper]{article}

\usepackage{cvpr}              
\usepackage{algorithm}
\usepackage{algorithmic}

\usepackage{newfloat}
\usepackage{listings}

\usepackage{amsmath}
\usepackage{graphicx}
\usepackage{multirow}
\usepackage{amssymb}
\usepackage{xcolor}
\usepackage{booktabs}

\usepackage{array}

\usepackage{times}
\usepackage{helvet}
\usepackage{courier}
\usepackage{xcolor}
\usepackage{marvosym}
\definecolor{cvprblue}{rgb}{0.21,0.49,0.74}
\usepackage[bookmarks=false,pagebackref,breaklinks,colorlinks,allcolors=cvprblue]{hyperref}

\makeatletter
\def\ps@myheadings{%
	\let\@oddfoot\@empty\let\@evenfoot\@empty
	\def\@evenhead{\thepage\hfil\slshape\leftmark}%
	\def\@oddhead{{\slshape\rightmark}\hfil\thepage}%
	\let\@mkboth\@gobbletwo
	\let\sectionmark\@gobble
	\let\subsectionmark\@gobble
}
\if@titlepage
\renewcommand\maketitle{\begin{titlepage}%
		\let\footnotesize\small
		\let\footnoterule\relax
		\let \footnote \thanks
		\null\vfil
		\vskip 60\p@
		\begin{center}%
			{\LARGE \@title \par}%
			\vskip 3em%
			{\large
				\lineskip .75em%
				\begin{tabular}[t]{c}%
					\@author
				\end{tabular}\par}%
			\vskip 1.5em%
			{\large \@date \par}
		\end{center}\par
		\@thanks\@notice
		\vfil\null
	\end{titlepage}%
	\setcounter{footnote}{0}%
}
\else
\renewcommand{\maketitle}{\par
	\begingroup
	\renewcommand\thefootnote{\@fnsymbol\c@footnote}%
	\def\@makefnmark{\rlap{\@textsuperscript{\normalfont\color{black}\@thefnmark}}}%
	\long\def\@makefntext##1{\parindent 1em\noindent
		\hb@xt@1.8em{%
			\hss\@textsuperscript{\normalfont\@thefnmark}}##1}%
	\if@twocolumn
	\ifnum \col@number=\@ne
	\@maketitle
	\else
	\twocolumn[\@maketitle]%
	\fi
	\else
	\newpage
	\global\@topnum\z@   
	\@maketitle
	\fi
	\thispagestyle{plain}\@thanks
	\endgroup
	\setcounter{footnote}{0}%
}
\makeatother

\usepackage{multibib}
\newcites{latex}{References}

\usepackage{tikz}
\usepackage{xcolor}
\usetikzlibrary{shapes.geometric}

\newcommand{\kidneypicture}{%
\begin{tikzpicture}[scale=0.5, line cap=round, line join=round]

\definecolor{darkpurple}{RGB}{68,30,110}
\fill[darkpurple]
  plot[smooth cycle, tension=1.0] coordinates {
    (-1.95,  0.10)
    (-1.55,  0.95)
    (-0.65,  1.20)
    ( 0.55,  1.00)
    ( 1.30,  0.45)
    ( 1.15, -0.05)
    ( 0.55, -0.28)
    ( 0.05, -0.10)
    ( 0.30, -0.78)
    (-0.45, -1.08)
    (-1.30, -0.88)
    (-1.82, -0.28)
  };

\foreach \x/\y in {
  -1.45/ 0.42,
  -1.10/ 0.82,
  -0.45/ 0.93,
   0.35/ 0.78,
   0.95/ 0.38,
  -1.18/-0.28,
  -0.65/-0.72,
   0.05/-0.58,
  -0.18/ 0.05
}{
  \node[
    star,
    star points=5,
    star point ratio=2.25,
    fill=green!70!black,
    draw=white,
    line width=0.35pt,
    minimum size=5pt,
    inner sep=0pt
  ] at (\x,\y) {};
}

\foreach \x/\y in {
  -2.20/ 0.10,
  -1.78/ 1.12,
  -0.70/ 1.42,
   0.55/ 1.20,
   1.48/ 0.52,
   1.38/-0.12,
   0.62/-0.98,
  -0.52/-1.28,
  -1.42/-1.02,
  -2.02/-0.35
}{
  \node[
    star,
    star points=5,
    star point ratio=2.25,
    fill=red!85!black,
    draw=white,
    line width=0.35pt,
    minimum size=6pt,
    inner sep=0pt
  ] at (\x,\y) {};
}

\end{tikzpicture}%
}

\def\paperID{****} 
\def\confName{CVPR}
\def\confYear{2026}

\usepackage{utfsym} 
\newcommand{\heart}{$\;\!$\usym{2665}}

\title{\begin{minipage}{0.88\textwidth}\centering
Focus on Background: Exploring SAM's Potential in Few-shot Medical Image Segmentation with Background-centric Prompting
\end{minipage}
\hspace{-0.1cm}
\begin{minipage}{0.11\textwidth}
\vspace{-0.05cm}
\kidneypicture
\end{minipage}
\vspace{-0.3cm}
}

\author{%
Yuntian Bo\textsuperscript{1} \quad
Yazhou Zhu\textsuperscript{1} \quad
Piotr Koniusz\textsuperscript{2,3,\Letter} \quad
Haofeng Zhang\textsuperscript{1,\Letter}\\
\textsuperscript{1}Nanjing University of Science and Technology \quad
\textsuperscript{2}University of New South Wales \quad
\textsuperscript{3}Data61$\!${\color{red}\heart}CSIRO \\
\vspace{-0.2cm}
{\tt\small \{yuntian.bo, zyz\_nj, zhanghf\}@njust.edu.cn, piotr.koniusz@unsw.edu.au}
}
\begin{document}

\maketitle

\begingroup
\renewcommand\thefootnote{}
\footnotetext{ \hspace{-3ex} \Letter\ Corresponding authors.$\qquad$This paper is accepted by CVPR'26.}
\endgroup

\begin{abstract}
Conventional few-shot medical image segmentation (FSMIS) approaches face performance bottlenecks that hinder broader clinical applicability. Although the Segment Anything Model (SAM) exhibits strong category-agnostic segmentation capabilities, its direct application to medical images often leads to over-segmentation due to ambiguous anatomical boundaries. In this paper, we reformulate SAM-based FSMIS as a prompt localization task and propose FoB (Focus on Background), a background-centric prompt generator that provides accurate background prompts to constrain SAM’s over-segmentation. Specifically, FoB bridges the gap between segmentation and prompt localization by category-agnostic generation of support background prompts and localizing them directly in the query image. To address the challenge of prompt localization for novel categories, FoB models rich contextual information to capture foreground-background spatial dependencies. Moreover, inspired by the inherent structural patterns of background prompts in medical images, FoB models this structure as a constraint to progressively refine background prompt predictions. Experiments on three diverse medical image datasets demonstrate that FoB outperforms other baselines by large margins, achieving state-of-the-art performance on FSMIS, and exhibiting strong cross-domain generalization. 
Our code is available at \href{https://github.com/primebo1/FoB_SAM}{https://github.com/primebo1/FoB\_SAM}.
\end{abstract}



\section{Introduction}
Few-shot medical image segmentation (FSMIS) \cite{ssl-alp,RPNET,ADNET} has achieved remarkable progress, showcasing its potential to reduce per-task labeling costs while maintaining accurate segmentation of anatomical structures and focal regions.
Despite recent advances, current FSMIS models face performance bottlenecks due to non-robust architectures and reliance on pseudo-label training, which limits generalization and compromises segmentation accuracy. Nonetheless, achieving high-precision automatic segmentation remains fundamental to dependable computer-aided diagnosis and treatment support in clinical practice.

\begin{figure}[t]
    \centering
    \includegraphics[width=0.99\linewidth]{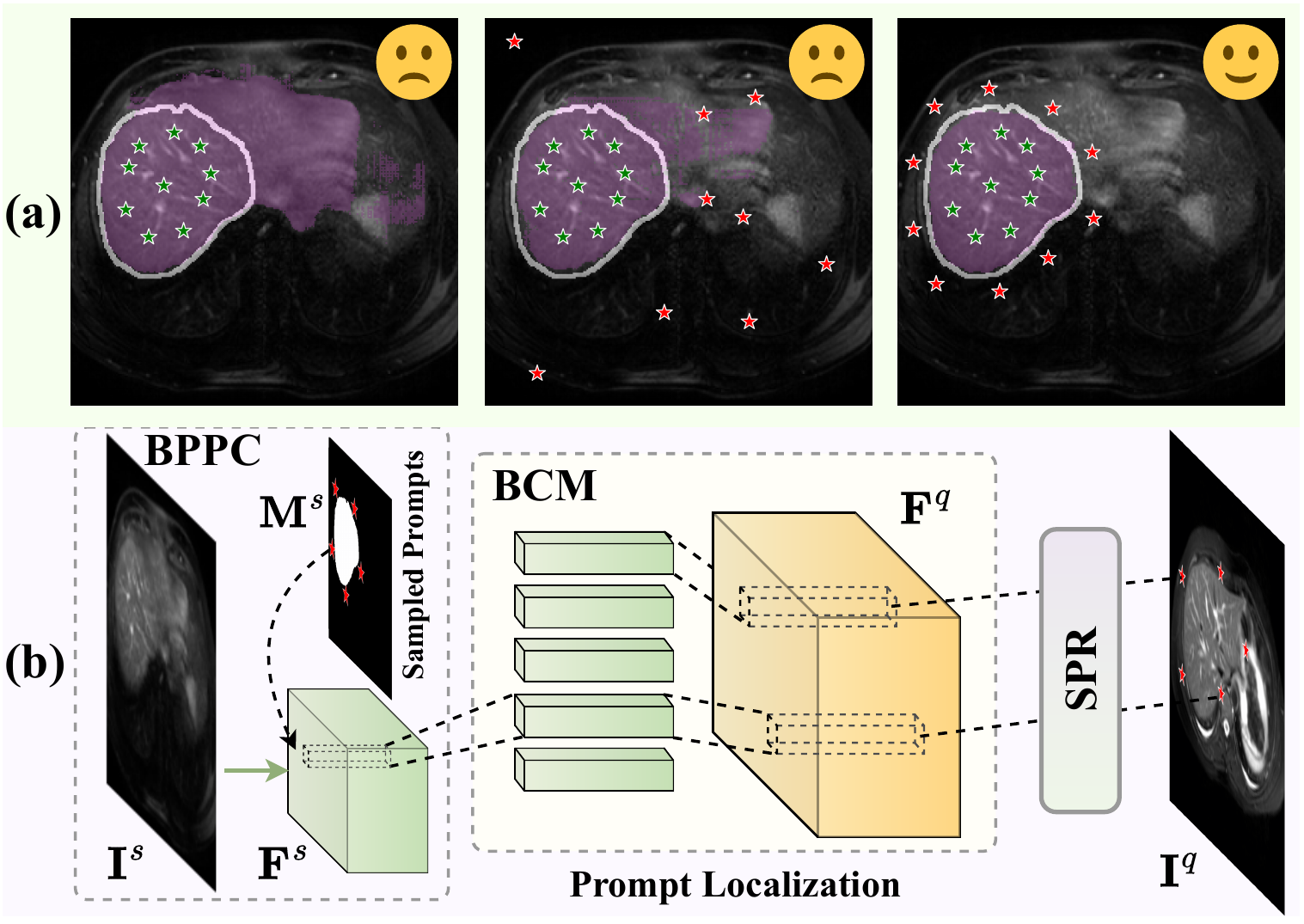}
    \caption{(a) Methodological motivation.
We observe that without accurate background prompts, SAM tends to over-segment ambiguous boundaries, highlighting the necessity of precise background guidance. 
(b) Technical motivation.
Prior methods extract prompts from a foreground segment map, yielding only reliable foreground prompts. We reformulate SAM-based FSMIS as a prompt localization task, focusing specifically on precise background prompt identification to limit SAM’s over-segmentation.
}
    \label{fig:motivation}
\end{figure}

The recently introduced foundation vision model, SAM \cite{Sam},  can improve FSMIS performance with minimal modifications. Trained on billion-scale segmentation tasks, SAM possesses promptable and category-agnostic segmentation capabilities, which naturally align with the goals of FSMIS. A recent study on incorporating SAM into FSMIS, ProtoSAM \cite{protosam}, demonstrates a successful case. It adopts a straightforward approach--it performs coarse segmentation using an FSMIS model, followed by selecting high-confidence points from the predicted probability map as prompts to guide SAM for refined segmentation.

However,  ProtoSAM \cite{protosam} remains suboptimal, in stark contrast to SAM’s impressive segmentation capability on natural images. To investigate the underlying cause, we analyze SAM’s segmentation behavior and observe a key issue: as shown in Figure \ref{fig:motivation}(a), SAM frequently \textbf{over-segments} medical images. We attribute this behavior to the limited ability of SAM, which is trained on natural images, to distinguish the ambiguous low-contrast boundaries prevalent in medical imaging, particularly at the junctions of adjacent organs or tissues \cite{AMBIGUITY}. Moreover, our findings reveal that \textbf{precise background prompts can effectively constrain over-segmentation.} However, ProtoSAM only provides accurate foreground prompts, overlooking the necessity of identifying reliable background prompts.

Based on the above insights, we argue that for SAM-based models, \textit{the  segmentation quality can be improved by improving the quality of background prompts rather than foreground prompts}. Thus, instead of approaching prompt generation from a segmentation perspective, we reformulate it as a direct background-centric point localization problem, and identify two key challenges: 1) \textbf{Reformulating segmentation for prompt localization.} The support segmentation reference needs to be transformed into appropriate descriptors that  effectively guide query prompt localization. 2) \textbf{Reliable prompts for novel categories.} Precise localization of a point for novel categories is extremely difficult, considering most background points lack semantic meaning. Moreover, reliable background prompts should be close to the category boundary to provide effective and targeted constraints.

To address the aforementioned challenges, this paper tackles the FSMIS task from a new perspective by constructing a plug-and-play prompt generator, termed \textbf{F}ocus \textbf{o}n \textbf{B}ackground (FoB). Inspired by few-shot landmark detection \cite{9878708,fsld1,fsld2,fsld3,fsld4,Lu_Koniusz_2024,10.1007/978-3-031-72655-2_9,10.1007/s11263-025-02671-5}, we directly guide FoB to localize background points near the object boundary, providing effective constraints for SAM-based segmentation.
To this end, a Background Prompt Prototype Construction (BPPC) module is introduced to extract background prompt locations from the support segmentation mask and form multiple background prompt prototypes for subsequent matching. Since valid background prompts typically surround the foreground, we propose Background-centric Context Modeling (BCM) to capture rich contextual interactions between the foreground and background (somewhat related to \cite{Koniusz09,10.1109/ICPR.2010.192,koniusz2011spatial,10.1007/s11263-025-02671-5}), as well as among background prompts, thereby compensating for the absence of fixed semantic patterns in the background regions surrounding novel categories.
Built upon the coarse predictions from BCM, we further introduce a Structure-guided Prompt Refinement (SPR) module to model structural dependencies among background prompts and refine the predictions accordingly, ensuring consistency with the expected spatial distribution of prompts for medical images.

Our method outperforms state-of-the-art (SOTA) approaches by a large margin, demonstrating superior generalization across three public datasets with diverse imaging modalities and anatomical regions. Furthermore, we observe that FoB maintains consistently strong performance under the challenging cross-domain setting.

\vspace{0.1cm}
In summary, our contributions are as follows:

\renewcommand{\labelenumi}{\roman{enumi}.}
\begin{enumerate}[leftmargin=0.6cm]
    \item  We reformulate SAM-based FSMIS as a prompt localization problem and design a dedicated prompt generator FoB, which focuses on background prompts.
    \item Our method effectively exploits the support segmentation information and leverages the contextual dependencies of medical image features to enable precise background prompt localization through matching.
    \item Our method leverages the structural dependencies among background prompts as constraints to refine the coarse predictions and correct poor prompt locations.
\end{enumerate}

\begin{figure*}[t]
    \centering
    \vspace{-2ex}
    \includegraphics[width=0.99\linewidth]{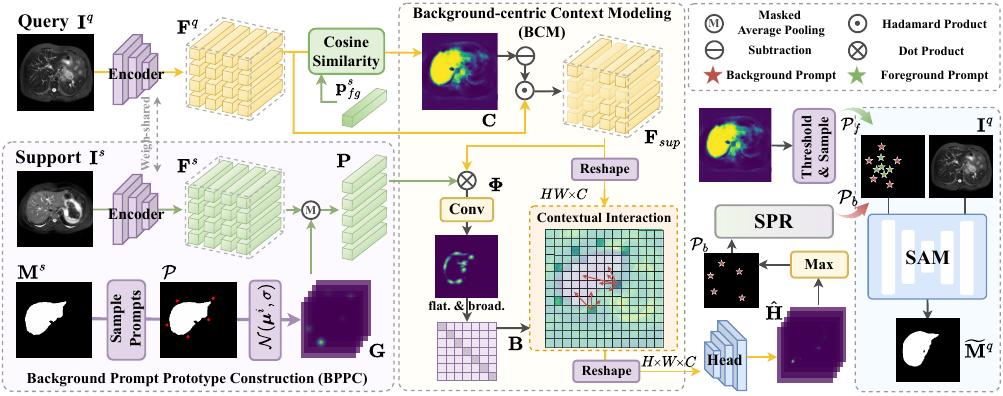}
    \vspace{-1ex}
    \caption{The overall architecture of the proposed FoB model for FSMIS, which consists of three collaboratively designed modules. BPPC prepares prompt prototypes for subsequent matching and localization. BCM models the dependency between background prompts and the foreground to produce coarse localization. SPR (see Figure \ref{fig:SPR_module}) imposes structural constraints in the feature space to refine background prompt predictions from BCM. FoB is trained independently of SAM and serves as a plug-and-play prompt generator during inference.}
    \label{fig:overall architecture}
    \vspace{-1ex}
\end{figure*}

\section{Related Works}
\subsection{Few-shot Medical Image Segmentation}

The key strength of FSMIS models \cite{ssl-alp,RPNET,SENET,GCN-DE,MRrNet} lies in their class-agnostic segmentation capability after training, letting them segment unseen categories with only a few annotated examples, thus reducing data dependence.
Currently, approaches based on prototypical networks \cite{qnet,CATNET,rpt,GMRD,PGRNET,PCMR,DSPNET,DIFD,COW} dominate FSMIS. Similarly to non-medical few-shot segmentation \cite{PANET,kang2023distilling}, these methods typically average features from support pixels belonging to the target class to form one or more prototypes, which are then compared with query features to identify the corresponding target regions.
Recent works focus on enhancing generalization by expanding training task diversity and improving support-query matching. For instance, \cite{ssl-alp} utilizes super-pixel pseudo-labels to diversify training tasks, while \cite{ADNET} further refines this with super-voxel labels tailored for 3D volumes. Huang \etal \cite{PGRNET} incorporate graph-based reasoning to jointly propagate support information while preserving query context. Tang \etal \cite{DSPNET} propose high-fidelity prototypes that preserve both semantic and structural cues for improved matching accuracy.
We also employ the prototypical setting but instead of matching dense segmentation masks, we tackle challenging  matching of sparse pixel features for prompt generation.

\subsection{Segment Anything Model for FSMIS}
Recent studies have explored integrating SAM into FSMIS, either by incorporating self-prompting intermediate modules \cite{MICCAI_w_SAM}, by introducing adapter structures \cite{WACV_SAM}, or by combining both, as in the concurrent work AM-SAM \cite{UnleashingSAM}. However, these methods require SAM to participate in or be fine-tuned during training, which results in high computational cost and low training efficiency.
To address this, ProtoSAM \cite{protosam} proposes a modular solution that connects a pre-trained FSMIS model to SAM by extracting prompts from coarse segmentation predictions, enabling independent training of the FSMIS module and reducing computational overhead. However, all existing methods focus solely on generating accurate foreground prompts, whereas our observations reveal that precise background prompts are indispensable for achieving accurate segmentation.
Our proposed FoB is tailored to directly generate prompts, with a particular focus on the accuracy of background prompts. It can be trained independently and serves as an efficient plug-and-play prompt generator for SAM during inference.

\section{Methodology}
\subsection{Problem Definition}
The conventional FSMIS task aims to meta-train a category-agnostic segmentation model $\text{Seg}(\cdot)$ on base categories $\mathcal{C}_b$ to enable rapid adaptation to novel categories $\mathcal{C}_n$ with only a few annotated references, where $\mathcal{C}_b \cap \mathcal{C}_n = \emptyset$. Specifically, during training, each meta-task is constructed as an episode $(\mathcal{S}, \mathcal{Q})$, where $\mathcal{S} = \big\{ (\mathbf{I}^s_i, \mathbf{M}^s_i) \big\}_{i=1}^K$ denotes the support set, $\mathcal{Q} = \big\{ (\mathbf{I}^q, \mathbf{M}^q) \big\}$ is the query set, and $\mathbf{I}$ and $\mathbf{M}$ denote the image and the mask, respectively. Most prior works adopt the challenging 1-shot setting ($K = 1$), which is also employed in this study. During inference, $\text{Seg}(\cdot)$ is meta-tested to segment unseen categories directly.

Distinct from conventional methods, we reformulate FSMIS as a prompt generation task and adopt the SAM model $\text{SAM}(\cdot)$ as the segmentation backend. Given the support set $\mathcal{S}$, we aim to enrich the available segmentation annotations into informative prompts, and automatically generate a prompt set $\boldsymbol\phi$ for each $\mathbf{I}^q \in \mathcal{Q}$. The final segmentation is then performed by feeding $\boldsymbol{\phi}$ and $\mathbf{I}^q$ into $\text{SAM}(\cdot)$.

\subsection{Overall Architecture}
The  proposed FoB from Figure \ref{fig:overall architecture} includes three key stages. 1) Background prompt prototypes are generated from the segmentation reference in the support set through Background Prompt Prototypes Construction (BPPC). 2) Rich contextual information between background prompts and foreground is captured through Background-centric Context Modeling (BCM) to facilitate matching between prompt prototypes and the query feature representation, resulting in a set of coarse query background prompt predictions.
3) The coarse predictions are refined through Structure-guided Prompt Refinement (SPR), which promotes feature-level support-query Structure Propagation with Graph (SPG) and performs Iterative Deformable Refinement (IDR) of prompt coordinates using the updated features.
The proposed model can be trained independently and serves as a plug-and-play prompt generator for SAM during inference.

\subsection{Background Prompt Prototypes Construction}
\label{section:BPPC}
In this stage, to bridge the gap between segmentation and point localization, the most suitable points from the support mask are sampled to generate corresponding prompt prototypes, which serve as the foundation for locating background prompts in the query image. To begin, we employ a weight-shared encoder $f(\cdot)$ to extract the support and query features: $\mathbf{F}^s = f(\mathbf{I}^s)$ and $\mathbf{F}^q = f(\mathbf{I}^q) \in \mathbb{R}^{H \times W \times C}$. Then, we obtain background prompts for $\mathbf{I}^s$ via $\mathbf{M}^s$:
\begin{equation}
\label{eq:sample_prompts}
\mathcal{P} = \mathcal{U}\big(\rho(\mathbf{M}^s, r) - \rho(\mathbf{M}^s, r-\epsilon), N_p\big), 
\end{equation}
where $\mathcal{P} \subset \mathbb{R}^2$  denotes the sampled prompts, $\mathcal{U}(\mathcal{R},n)$ denotes a uniform sampling function that samples $n$ points from region $\mathcal{R}$, $\rho(\mathbf{M}, r)$ denotes a dilation function on mask $\mathbf{M}$ given a dilation kernel of size $r=15$. We set $\epsilon=2$ to create a ``differential region'' between the two dilated masks which enables uniform sampling of $N_p$ points from the ``differential region'' between the two differently dilated areas.

For each point $\boldsymbol{\mu}^i\in\mathcal{P}$, following \cite{HRNET}, a Gaussian heatmap centered at $\boldsymbol{\mu}^i$ is generated to construct the heatmap set $\mathbf{G} \in \mathbb{R}^{N_p \times H \times W}$, formalized as:
\begin{equation}
\label{heatmap}
\mathbf{G} = [\mathbf{G}^1, \mathbf{G}^2, \dots, \mathbf{G}^{N_p}], \quad \mathbf{G}^i = \mathcal{N}(\boldsymbol{\mu}^i, \sigma),
\end{equation}
where $\mathcal{N}(\boldsymbol{\mu}^i, \sigma)$ denotes a 2D Gaussian distribution with center $\boldsymbol{\mu}^i$ and a standard deviation $\sigma$. The heatmaps $\mathbf{G}^i$ are concatenated along the channel dimension to form  $\mathbf{G}$.

Subsequently, we use Masked Average Pooling (MAP) to construct the background prompt prototype set 
$
\mathbf{P} = \big[\mathbf{p}^1_{b}, \mathbf{p}^2_{b},..., \mathbf{p}^{N_p}_{b}\big] \in \mathbb{R}^{N_p \times C}
$ as follows:
\begin{equation}
\label{MAP}
\mathbf{p}^i_{b} = \operatorname{MAP}(\mathbf{F}^s, \mathbf{G}^i) = \frac{\sum_{u, v} \mathbf{F}^s{(u, v)}\mathbf{G}^i{(u, v)}}{\sum_{u, v}  \mathbf{G}^i{(u, v)}}, i \in [1, N_p],
\end{equation}
where $\mathbf{p}^i_{b} \in \mathbb{R}^{C}$ denotes the $i$-th background prompt prototype of $\mathbf{P}$, $\mathbf{F}^s{(u, v)}\in\mathbb{R}^C$ and $\mathbf{G}^i{(u, v)}$ is a scalar at any given  $(u,v)$. MAP uses heatmaps as weighting maps to compute local weighted averages, enabling accurate point vector extraction and aliasing reduction.

\subsection{Background-centric Context Modeling}
A key challenge in background prompt localization is the lack of explicit semantic patterns, due to the unstructured and non-semantic nature of background regions in medical images.
We consider contextual information, \ie, the spatial layout and relative relations of background prompts around the foreground, as a learning objective, which remains valid even for novel categories during inference.

To this end, BCM operates in a coarse-to-fine manner, guiding the model to learn prompt context in a background-centric manner. To begin, we suppress the foreground region as an initial step to facilitate background–foreground differentiation in the subsequent modeling steps:
\begin{equation}
\label{eq4}
    \mathbf{F}_{sup} = \bigg( 1 - \frac{\mathbf{F}^q \cdot \mathbf{p}^s_{fg}}{\|\mathbf{F}^q\|_2 \|\mathbf{p}_s^{fg}\|_2} \bigg) \odot \mathbf{F}^q = ( 1 - \mathbf{C} ) \odot \mathbf{F}^q,
\end{equation}
where $\mathbf{F}_{sup} \in \mathbb{R}^{H \times W \times C} $ is the foreground-suppressed feature map, $\mathbf{p}^s_{fg} = \operatorname{MAP}(\mathbf{F}^s, \mathbf{M}^s)$ denotes the support foreground prototype, $\odot$ denotes the Hadamard product, and $\mathbf{C} \in \mathbb{R}^{H \times W}$ denotes the correlation map.

Subsequently, we generate coarse prompt proposals to indicate background prompt locations for the following steps as:
\begin{equation}
\label{eq5}
    \mathbf{\Phi} = \xi^{-1}\big((\mathbf{A} \odot\mathbf{P}\mathbf{W}_s)
    ( \mathbf{W}_q\,\xi(\mathbf{F}_{sup}))
    \big),
\end{equation}
where $\mathbf{\Phi} \in \mathbb{R}^{N_p \times H \times W}$ denotes a tensor with each channel representing a prompt location proposal. $\mathbf{W}_s,\mathbf{W}_q \in \mathbb{R}^{C \times C}$ are learnable projection matrices, $\mathbf{A} \in \mathbb{R}^{N_p \times C}$ from \cite{dynamic_sim} is a channel attention weight conditioned on $\mathbf{P}$ used to distinguish different $\mathbf{p}_b^i$ of $\mathbf{P}$, and $\xi: \mathbb{R}^{d \times H \times W} \to \mathbb{R}^{d \times HW}$ reshapes the spatial dimensions independently of $d$, with $\xi^{-1}$ as its inverse.

Next, the suppressed feature map tensor $\mathbf{F}_{sup}$ is fed into a masked attention transformer to model global contextual dependencies via pixel-wise feature interaction, where the proposal tensor $\mathbf{\Phi}$ serves as a soft mask that assigns higher attention weights to prompt locations through so-called bias  $\mathbf{B} = \operatorname{ReLU}(\mathcal{C}(\Phi))$:
\begin{align}
\label{eq6}
\mathbf{F}_{sup}' &= \operatorname{LN}\big(\operatorname{MHA}(\xi(\mathbf{F}_{sup}); \mathbf{B}) + \xi(\mathbf{F}_{sup})\big), \\
\label{eq7}
\mathbf{F}_{m} &= \xi^{-1}\big(\operatorname{LN}(\operatorname{FFN}(\mathbf{F}_{sup}') + \mathbf{F}_{sup}')\big),
\end{align}
where $\mathbf{F}_{m} \in \mathbb{R}^{C \times H \times W}$ denotes the modulated feature tensor. $\operatorname{LN}(\cdot)$ is the layer normalization function, and $\operatorname{FFN}(\cdot)$ is the feed-forward network. $\operatorname{MHA}(\cdot; \mathbf{B})$ refers to multi-head self-attention with a bias matrix added to the attention logits to modulate attentional focus. The bias $\mathbf{B}$, 
 defined above Eq. \eqref{eq6}, uses
 a $1 \times 1$ convolution  $\mathcal{C}(\cdot)$  to compress channel-wise information into a single attention map. This map is then flattened and broadcast to construct the bias matrix, thereby emphasizing features in the coarsely activated background prompt regions. Given the numerically separated foreground and background in $\mathbf{F}_{sup}$ and the spatial indication of background prompts in $\mathbf{\Phi}$, the transformer can readily model their relative relations through pairwise interactions between pixel-level features.

Finally, $\mathbf{F}_m$ is an input to a lightweight detection head $\text{Head}(\cdot)$ to obtain the background prompt heatmap $\hat{\mathbf{H}} = \text{Head}(\mathbf{F}_m) \in \mathbb{R}^{N_p \times H \times W}$, where each channel represents a prompt prediction. A set of coarse background prompt coordinates $\mathcal{P}_b = \big\{\boldsymbol{\mu}_b^1, \boldsymbol{\mu}_b^2, \dots, \boldsymbol{\mu}_b^{N_p} \big\}$ is obtained by selecting the point with the maximum response in each channel of $\hat{\mathbf{H}}$.

\subsection{Structure-guided Prompt Refinement}
Background prompts in medical images often exhibit a highly structured spatial distribution, typically forming a ring-like shape around the target object when sequentially connected. However, in the output $\mathcal{P}_b$ from the previous stage, we frequently observe outlier prompts that deviate from this regular pattern, \eg, prompts may collapse into a compact cluster or split into multiple disconnected groups. We attribute this to the lack of constraints on the geometric relationships between prompts in earlier stages, which treat them as independent entities. Therefore, we introduce structural priors as constraints to calibrate the geometric distribution of predicted prompts.

\vspace{0.1cm}
\noindent \textbf{Structure Propagation with Graph.} Our goal is to enable the query prompts to ``perceive'' the structural patterns encoded in the support features, thereby facilitating structure-aware refinement. Since graph structures are inherently suitable for modeling such relationships, we construct a graph that explicitly encodes the support structures and propagates the encoded signals to the query through cross-instance graph convolution (GCN). This allows the query representation to align with the support structure through graph message passing. Considering the distributional variations of background prompts across different categories, we first adaptively estimate the graph structure $\mathbf{A}^{ada}$:
\begin{equation}
\label{eq8}
   \mathbf{A}^{ada}
   =\operatorname{softmax}\Big(
     \tfrac{1}{\sqrt{C}}\,
\mathbf{P}\mathbf{W}_{\theta}\bigl(\mathbf{P}\mathbf{W}_{\phi}\bigr)^{\top} 
   \Big) \in \mathbb{R}^{N_p \times N_p},
\end{equation}
where $\mathbf{W}_{\theta},\mathbf{W}_{\phi} \in \mathbb{R}^{C \times C}$ are two  projection matrices, and $\tfrac{1}{\sqrt{C}}$ is a scaling factor to ensure numerical stability. 

Then, for the ring prior, we define a static structure $\mathbf{A}^{ring} \in \mathbb{R}^{N_p \times N_p}$ where each prompt exchanges messages with its two adjacent prompts along the ring to smooth features and prevent the emergence of outlier representations:
\begin{equation}
\label{eq9}
(\mathbf{A}^{ring})_{ij}
=\begin{cases}
1,& j=(i\pm1) \pmod{N_p},\\
0,& \text{otherwise}.
\end{cases}
\end{equation}

\begin{figure}
    \centering
    \vspace{-1ex}
    \includegraphics[width=0.99\linewidth]{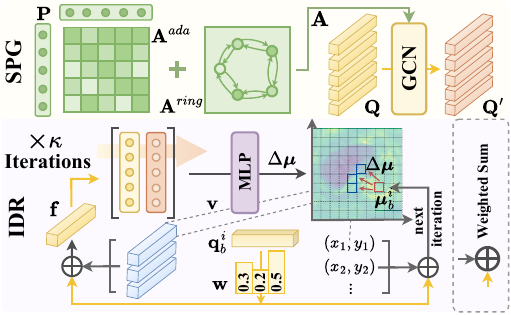}
    \vspace{-1ex}
    \caption{Schema of Structure-guided Prompt Refinement.}
    \label{fig:SPR_module}
    \vspace{-2ex}
\end{figure}

Afterwards, the structural representation $\mathbf{A}$ of the support background prompts is computed as a weighted sum of the adaptive and ring structures, which is formalized as:
\begin{equation}
\label{eq10}
\mathbf{A}=\alpha\,\mathbf{A}^{ada}
+(1-\alpha)\,\mathbf{A}^{ring},
\end{equation}
where $\alpha \in [0,1]$ is a learnable weight factor.

Finally, the support-conditioned structure is used to update the query prompt prototypes via a GCN, allowing the structural prior from the support to constrain the query background prompt prototypes $\mathbf{Q} = \big[\mathbf{q}^1_{b}, \mathbf{q}^2_{b},..., \mathbf{q}^{N_p}_{b}\big] \in \mathbb{R}^{N_p \times C}$, which are obtained same way as  in Eq.~\eqref{heatmap} and \eqref{MAP} using the predicted prompts $\mathcal{P}_b$:
\begin{equation}
\label{eq11}
   \mathbf{Q}'
   =\operatorname{ReLU}\big(
      \mathbf{D}^{-\frac{1}{2}}
      \mathbf{A}
      \mathbf{D}^{-\frac{1}{2}}
      \mathbf{Q}\mathbf{W}_{g} 
   \big),
\end{equation}
where $\mathbf{Q}' = \big[{\mathbf{q}^1_{b}}', {\mathbf{q}^2_{b}}',..., {\mathbf{q}^{N_p}_{b}}'\big] \in  \mathbb{R}^{N_p \times C}$ denotes the updated query prompt prototypes, $\mathbf{D}$ is the degree matrix of $\mathbf{A}$ used for normalization, and $\mathbf{W}_{g} \in \mathbb{R}^{C \times C}$ denotes a learnable projection matrix.

\vspace{0.1cm}
\noindent \textbf{Iterative Deformable Refinement.}
The previous operations calibrate the distribution of query prompt features in the feature space to better fit the inherent structure of background prompts, yet the predicted prompt coordinates in $\mathcal{P}_b$ remain unchanged. Inspired by deformable attention \cite{deformable_attention}, we propose to leverage the updated query features to guide prompt location refinement in an iterative deformable manner.
Specifically, for each $\boldsymbol{\mu}_b^i \in \mathcal{P}_b$ and its corresponding prototype $\mathbf{q}_b^i \in \mathbf{Q}$, a direction vector $\mathbf{v}$ is formed to encode the direction and magnitude of feature displacement, which is then used to predict offsets $\Delta\boldsymbol{\mu}_m\in \mathbb{R}^{2}$ for $m=1,\ldots,k$:
\begin{equation}
\label{eq12}
\big[\Delta\boldsymbol{\mu}_1,\cdots,\Delta\boldsymbol{\mu}_k\big] = \phi(\mathbf{v}) = \phi\big(\big[\mathbf{q}_b^i, \mathbf{f}\big]\big),
\end{equation}
where $\mathbf{f}:={\mathbf{q}_b^i}'$ for the first iteration, $\phi: \mathbb{R}^{2C} \rightarrow \mathbb{R}^{k \times 2}$ is a two--layer fully connected network with ReLU for predicting $k$ spatial offsets, and $[\cdot,\cdot]$ denotes vector concatenation. 
For each point, we compute $k$ candidate offsets to obtain a smooth and flexible estimation of the refined coordinate. Thus, a set of weights $\mathbf{W}$ conditioned on the input features is computed to aggregate all candidates, formalized as:
\begin{equation}
\label{eq13}
\mathbf{w} =[w_1,\ldots,w_k]^\top=\operatorname{softmax}\big(\mathbf{q}_b^i \,\mathbf{W}_{att}\big) \in \mathbb{R}^k,
\end{equation}
where $\mathbf{W}_{att} \in \mathbb{R}^{C \times k}$ is a learnable projection matrix.

Finally, each coordinate $\boldsymbol{\mu}_b^{i}$ is refined as the weighted sum of $(\boldsymbol{\mu}_b^i + \Delta\boldsymbol{\mu}_m)$, and $\mathbf{f}$ is updated by aggregating query features at the corresponding  positions for the next iteration:
\begin{align}
\label{eq14}
    \boldsymbol{\mu}_b^{i} &= \sum_{m=1}^{k} w_m \left( \boldsymbol{\mu}_b^i + \Delta\boldsymbol{\mu}_{m} \right) \in \mathbb{R}^2, \\
    \label{eq15}
    \mathbf{f} &= \sum_{m=1}^{k} w_{m} \, \operatorname{bilinear} \left( \mathbf{F}_q, \boldsymbol{\mu}_b^i + \Delta\boldsymbol{\mu}_{m} \right) \in \mathbb{R}^C,
\end{align}
where $\operatorname{bilinear}(\mathbf{F}, \boldsymbol{\mu})$ denotes a bilinear sampler that extracts the feature vector at location $\boldsymbol{\mu}$ from the feature map tensor $\mathbf{F}$.

We repeat the  steps in Eq. \eqref{eq12}, \eqref{eq13}, \eqref{eq14} \& \eqref{eq15} for $\kappa$ iterations, during which the coordinates of prompts in $\mathcal{P}_b$ are progressively refined to approximate the locations consistent with the updated feature distribution $\mathbf{Q}'$, resulting in a refined background prompt set $\mathcal{P}'_b = \big\{{{\boldsymbol{\mu}'}_b^1}, {{\boldsymbol{\mu}'}_b^2}, \dots, {{\boldsymbol{\mu}'}_b^{N_p}} \Big\}$.

\subsection{Optimization}
\label{sec:optimization}
\noindent \textbf{Region-aware Contrast.} 
A major error occurs when the predicted background prompts mistakenly enter the foreground region. However, the distance-based regression loss functions only optimize predictions to approximate the foreground edges (where the support prompts are sampled), which may result in predictions falling into the foreground region due to inaccurate matching. Consequently, we propose a Region-aware Contrastive (RAC) Loss $\mathcal{L}_{rac}$ based on InfoNCE \cite{infoNCE} to differentiate the encodings of foreground and surrounding regions, thereby minimizing the risk of matching-based predictions falling into the foreground. This loss is formalized as:
\begin{equation}
\mathcal{L}_{rac} =  -\log\left(\frac{e^{\operatorname{sim}\big(\mathbf{p}^s_{fg}, \mathbf{p}^s_{p}\big) / \tau}}{\sum_{i=1}^{N_p} e^{\operatorname{sim}\big(\mathbf{p}^s_{fg}, \mathbf{p}^i_{b}\big) / \tau}}\right), 
\end{equation}
where $\operatorname{sim}(\cdot,\cdot)$ is the cosine similarity, $\mathbf{p}^s_{p}$ denotes the positive sample, and $\tau=0.1$ is the temperature. Notably, $\mathbf{p}^s_{p}$ is selected as the mean of the pixel features from the outermost region of the support foreground, ensuring consistency with the interior, and each $\mathbf{p}^i_{b}$ serves as a negative sample. This encourages the model to learn to distinguish between features inside and outside the boundary, thereby preventing incorrect matching.

\vspace{0.1cm}
\noindent \textbf{Prompt Regression.}
Following \cite{HRNET}, we compute the loss between the predicted coarse- and fine-grained background prompt heatmaps and their corresponding ground truth (GT) using the mean squared error (MSE) function:
\begin{equation}
\mathcal{L}_{heat} = \frac{1}{N_pHW} \left(
\big\|\mathbf{\Phi} - \mathbf{H}\big\|_F^2 + \big\|\hat{\mathbf{H}} - \mathbf{H}\big\|_F^2 \right) ,
\end{equation}
where $\mathbf{H}$ denotes the ground truth heatmaps computed using Eq. \eqref{heatmap} with the corresponding $\mathbf{M}^{q}$. Moreover, $\lVert\cdot\rVert_F$ in the above context means simply that tensors are reshaped into matrices and the Frobenius norm is applied.

To supervise SPR for better coordinate refinement, we also employ MSE to measure the error between the predicted coordinates and their corresponding GT:
\begin{equation}
    \mathcal{L}_{coor} = \frac{1}{N_p} \sum_{i=1}^{N_p} \big\| {{\boldsymbol{\mu}'}_b^i} - {\boldsymbol{\mu}}_q^i \big\|_2^2 ,
\end{equation}
where ${\boldsymbol{\mu}}_q^i$ denotes the $i$-th ground truth query background prompt coordinate, sampled as in Eq. \eqref{eq:sample_prompts}.

\vspace{0.1cm}
\noindent \textbf{Foreground Understanding.} To guide foreground prompt extraction and enhance foreground-background discrimination in BCM, we define a cross-entropy loss to optimize the pixel-wise classification probability distribution:
\begin{equation}
\mathcal{L}_{fore} = -\frac{1}{HW} \boldsymbol{1}^\top \big(\mathbf{M}^{q} \log(\mathbf{C}) + (1 - \mathbf{\mathbf{M}}^{q}) \log(1-\mathbf{C})\big)\boldsymbol{1},
\end{equation}
where $\boldsymbol{1}$ is all-ones vector. Overall, the total loss is defined as $\mathcal{L}_{total} = \mathcal{L}_{rac} + \lambda_1\mathcal{L}_{heat} + \lambda_2\mathcal{L}_{coor} +
\mathcal{L}_{fore}$.

\begin{table*}[t]
\centering
\caption{Comparison with SOTA methods (in Dice score \%) on Abd-MRI, Abd-CT, and Skin-DS under Settings I and II. The best values are shown in bold font. Indicator “--” means that the original paper did not report results or release the code for a fair comparison.}
        \vspace{-1ex}
\renewcommand\arraystretch{1.2}
\resizebox{\textwidth}{!}{
\begin{tabular}{c|r | c c c c c | c c c c c | c c c c } 
\toprule
\multicolumn{1}{c}{} &   \multirow{2}{*}{\textbf{Methods}}& \multicolumn{5}{c|}{\textbf{Abd-MRI}} & \multicolumn{5}{c|}{\textbf{Abd-CT}} & \multicolumn{4}{c}{\textbf{Skin-DS}} \\
\multicolumn{1}{c}{} & & Liv& RK & LK & Spl& Avg. & Liv& RK & LK & Spl& Avg. & Mel& Nev& SK & Avg. \\
\midrule
\multirow{8}{*}{\rotatebox{90}{Setting I}}& ALPNet \cite{ssl-alp} & 76.1& 85.18& 81.92& 72.18& 78.84&78.29& 71.81& 72.36& 70.96& 73.35& 66.32& 61.65& 59.57& 62.51\\
& RPT \cite{rpt} & 82.86& 89.82& 80.72& 76.37& 82.44& 82.57& 72.58& 77.05& 79.13& 77.83& 77.81& 75.42& 70.28& 74.50\\
& GMRD \cite{GMRD}& 81.42& {90.12}& 83.96& 76.09& 82.90& 79.6&  74.46& 81.7& 78.31& 78.52&   \textbf{79.23}& 72.78& 71.32& 74.44\\
& PGRNet \cite{PGRNET}& {83.27}& 87.44& 81.44& 81.72& 83.47&  82.48& 79.88& 74.23& 72.09& 77.17& 71.39& 70.21& 65.87& 69.16\\
& ProtoSAM \cite{protosam}& 83.14& 82.36& 82.75&  77.98& 81.56& 84.79& 75.67& 71.31& 70.24& 75.50& 73.61&  76.26& 68.37& 72.75\\
&AM-SAM \cite{UnleashingSAM}& 76.12&   84.95&   84.17&   \textbf{80.36}&   81.40& \textbf{87.28}&  86.01& 84.37& \textbf{87.11}& 86.19& -- & -- & -- & -- \\
&\textcolor{gray}{\textbf{FoB + S-2D}} &  \textcolor{gray}{77.09}& \textcolor{gray}{89.45}& \textcolor{gray}{83.58}& \textcolor{gray}{79.82}& \textcolor{gray}{82.49}& \textcolor{gray}{85.54}& \textcolor{gray}{80.02}& \textcolor{gray}{79.18}&  \textcolor{gray}{78.06}& \textcolor{gray}{80.70}& \textcolor{gray}{\textbf{85.87}}& \textcolor{gray}{\textbf{88.51}}& \textcolor{gray}{\textbf{80.02}}& \textcolor{gray}{\textbf{84.80}}\\
& \textbf{FoB + SAM}& \textbf{85.61}& \textbf{88.18}& \textbf{84.76}& {79.31}& \textbf{84.46}& {86.51}& {86.51}&  \textbf{87.29}& {84.54}& \textbf{86.21}& 78.93&  \textbf{77.12}&  \textbf{73.81}&  \textbf{76.62}\\
 \midrule
\multirow{7}{*}{\rotatebox{90}{Setting II}}& ALPNet \cite{ssl-alp}& 73.05& 78.39& 73.63& 67.02& 73.02& 73.67& 54.82& 63.34& 60.25& 63.02& 56.17& 50.67& 49.18& 52.01\\
& RPT \cite{rpt}& 76.37& 86.01& {78.33}& {75.46}& 79.04& 75.24& 67.73&   72.99&   70.8& 71.69& 76.07& 76.97& 69.86& 74.30\\
& GMRD \cite{GMRD}& 80.25& 86.66&\textbf{78.65}& 73.25& {79.70}& 80.39& 76.17& 77.4& 75.3&   77.32& \textbf{77.21}& 74.12& 70.97& 74.10\\
& ProtoSAM \cite{protosam}& {81.94}& 81.43& 71.46& \textbf{76.51}& 77.83&   \textbf{87.84}& 71.04& 69.44& 65.5& 73.45& 75.33& 72.01& 68.74& 72.03\\
&AM-SAM \cite{UnleashingSAM}& 79.70&   81.46&   70.28&   70.80&   75.56& 85.40&  84.02& 82.78& \textbf{83.97}& 84.04& -- & -- & -- & -- \\
&\textcolor{gray}{\textbf{FoB + S-2D}} & \textcolor{gray}{75.32}&   \textcolor{gray}{87.07}&   \textcolor{gray}{75.46}&   \textcolor{gray}{75.32}&   \textcolor{gray}{78.29}& \textcolor{gray}{75.25}&  \textcolor{gray}{78.97}& \textcolor{gray}{79.89}& \textcolor{gray}{75.82}& \textcolor{gray}{77.48}& \textcolor{gray}{\textbf{85.53}}& \textcolor{gray}{\textbf{87.02}}& \textcolor{gray}{\textbf{78.86}}& \textcolor{gray}{\textbf{83.80}}\\
& \textbf{FoB + SAM}&  \textbf{82.43}& \textbf{87.91}& 78.21& 73.30& \textbf{80.46}& {82.29}& \textbf{85.91}& \textbf{88.55}& {82.43}& \textbf{84.80}&  {76.68}&  \textbf{77.77}&  \textbf{72.22}&  \textbf{75.56}\\
\bottomrule
\end{tabular}

}
\label{table:main}
\end{table*}

\subsection{Inference}
Our proposed model can serve as a plug-and-play prompt generator during inference, providing both foreground and background prompt points for SAM. 
Notably, we uniformly sample $N_f$ points from $\mathbf{C}$ where the similarity values exceed a threshold $\mathcal{T}$ as foreground prompts, and $\mathcal{T}$ is empirically set to 0.9. The inference process is represented as:
\begin{equation}
\big(\mathcal{P}_f, \mathcal{P}'_b\big) = \text{FoB}\big(\mathbf{I}^s, \mathbf{M}^s, \mathbf{I}^q\big), \quad \widetilde{\mathbf{M}}^{q} = \text{SAM}\big(\mathbf{I}^q, \mathcal{P}_f, \mathcal{P}'_b\big),
\end{equation}
where $\mathcal{P}_f$ denotes the foreground prompts set, $\text{FoB}(\cdot)$ denotes the proposed FoB model, and $\widetilde{\mathbf{M}}^{q}$ is the final query mask prediction.


\section{Experiments}


\noindent \textbf{Datasets.} 
We comprehensively evaluate our model on three datasets with different image modalities and medical regions: Abd-CT \cite{abdct}, Abd-MRI \cite{abdmr}, and Skin-DS \cite{isic1,isic2}.
Abd-CT includes 30 3D CT scans and Abd-MRI contains 20 3D T2-SPIR MRI scans, and we selected their common four labels: liver (Liv), right kidney (RK), left kidney (LK), and spleen (Spl) for evaluation;
Skin-DS comprises 2594 dermoscopic skin lesion images, including 519 melanoma (Mel), 1867 nevus (Nev), and 208 seborrheic keratosis (SK) images, for assessment.
Appendix \ref{sup:superpix} details superpixel-based pseudo-labeling for Skin-DS.

\vspace{0.1cm}
\noindent \textbf{Implementation Details.} 
We use ResNet-101 \cite{RESNET} pre-trained on MS-COCO \cite{MSCOCO} as the feature encoder $f(\cdot)$ for our proposed model and comparison methods. All experiments are based on a 1-way 1-shot setting and all images are reshaped into 256$\times$256. For Abd-CT and Abd-MRI, we adopt the same pre-processing techniques as in \cite{ADNET} and \cite{ssl-alp}. The 3D supervoxel clustering method \cite{ADNET} is utilized to generate pseudo-masks for supervision during training, and the mean of 5-fold cross-validation results is reported. We adopt the experimental settings \cite{ssl-alp,ADNET} where in Setting I, test categories may appear unlabeled in the training image backgrounds, while in Setting II, test categories are entirely unseen during training. For Skin-DS, we propose Setting I, where training is performed using pseudo labels generated by SLIC superpixel \cite{slic} with all three diseases, and the experimental results are based on 5-fold cross-validation. In Setting II, two diseases are selected as seen categories for training with ground truths, while the third is reserved for testing. This process is rotated for cross-validation. We employ the `ViT-H' SAM during inference. 

Our model is implemented with PyTorch \cite{pytorch} and trained on an NVIDIA
RTX 4080S GPU for 36K iterations with a batch size of 1. We chose the Adam optimizer \cite{adam} with an initial learning rate of $1\times10^{-4}$ and a decay factor of 0.95 every 1K iterations.  The default number of prompts is set to $N_p = N_f = 10$. To balance the loss functions, $\lambda_1$ and $\lambda_2$ are set to $1\times10^{3}$ and $1\times10^{-4}$, respectively.
In SPR, we set the number of predicted offsets $k = 8$ and the number of refinement iterations $\kappa = 3$. To ensure a fair comparison, we employ the Dice coefficient \cite{ssl-alp} as the evaluation metric.

\begin{table}[t]
\centering
\vspace{-1ex}
\caption{Comparison with SOTA methods (in Dice score \%) under cross-domain setting using abdominal datasets. The best values are shown in bold font.}
        \vspace{-1ex}
\renewcommand\arraystretch{1.2}
\resizebox{0.95\linewidth}{!}{
\begin{tabular}{c|r |c c c c c  }
\toprule
\multicolumn{1}{c}{} & Methods& Liv& RK & LK & Spl& Avg. \\
\midrule
\multirow{3}{*}{\rotatebox{90}{\scriptsize CT $\to$ MRI}}& RobustEMD \cite{ROBUSTEMD} & 60.16& 70.26& 66.34& 53.71& 62.61\\
& FAMNet \cite{FAMNet}& 73.01& 74.68& 57.28& 58.21& 65.79\\
& \textbf{FoB + SAM}& \textbf{75.05}& \textbf{79.57}& \textbf{70.38}& \textbf{68.21}& \textbf{73.30}\\
\midrule
\multirow{3}{*}{\rotatebox{90}{\scriptsize MRI $\to$ CT}}& RobustEMD \cite{ROBUSTEMD} & 69.82& 50.34& \textbf{63.79}& 59.88& 60.95\\
& FAMNet \cite{FAMNet}& 73.57& \textbf{61.89}& 57.79& 65.78& 64.75\\
& \textbf{FoB + SAM}& \textbf{81.36}& 58.81&57.18& \textbf{70.71}& \textbf{67.02}\\
\bottomrule
\end{tabular}

}

\label{table:cross_domain}
\vspace{-2ex}
\end{table}

\subsection{Comparison with the State of the Art}
We compare our FoB model with current SOTA FSMIS models, including \cite{ssl-alp,rpt,GMRD,PGRNET,UnleashingSAM,protosam}. As shown in Table \ref{table:main}, our method consistently outperforms on Abd-CT and Abd-MRI. Especially, the results on Abd-CT under Setting I and Setting II achieves 86.21\% and 84.80\%, respectively, outperforming the previous best conventional FSMIS method \cite{GMRD} by 7.69\% and 7.48\%. Significant improvements are also observed on the Abd-MRI and Skin-DS datasets, where Dice scores are on average 1.16\% and 1.69\% higher than the second-best method across two different settings, respectively. To assess FoB's generalization, we validate it with a SOTA medical SAM, SAM-Med2D (S-2D) \cite{SAMMED2D}, as shown in the gray rows of the table. As shown in Table \ref{table:main}, FoB+S-2D slightly underperforms FoB+SAM on abdominal datasets but significantly surpasses it on Skin-DS. However, using medical SAM variants may violate FSMIS protocol, we compare this variant for assessing generalization rather than performance boosting. We provide a detailed discussion in the Appendix. Moreover, our method also  outperforms previous methods based on SAM \cite{protosam,UnleashingSAM}. A concurrent work, AM-SAM \cite{UnleashingSAM}, achieves comparable performance to ours on Abd-CT. However, its results on Abd-MRI are significantly inferior due to the blurred boundaries in MRI images. 
Moreover, AM-SAM \cite{UnleashingSAM} jointly fine-tunes the SAM model and trains the prompt generator, substantially increasing computational cost.

\begin{figure}[t]
    \centering
    \vspace{-1ex}
    \includegraphics[width=0.9\linewidth]{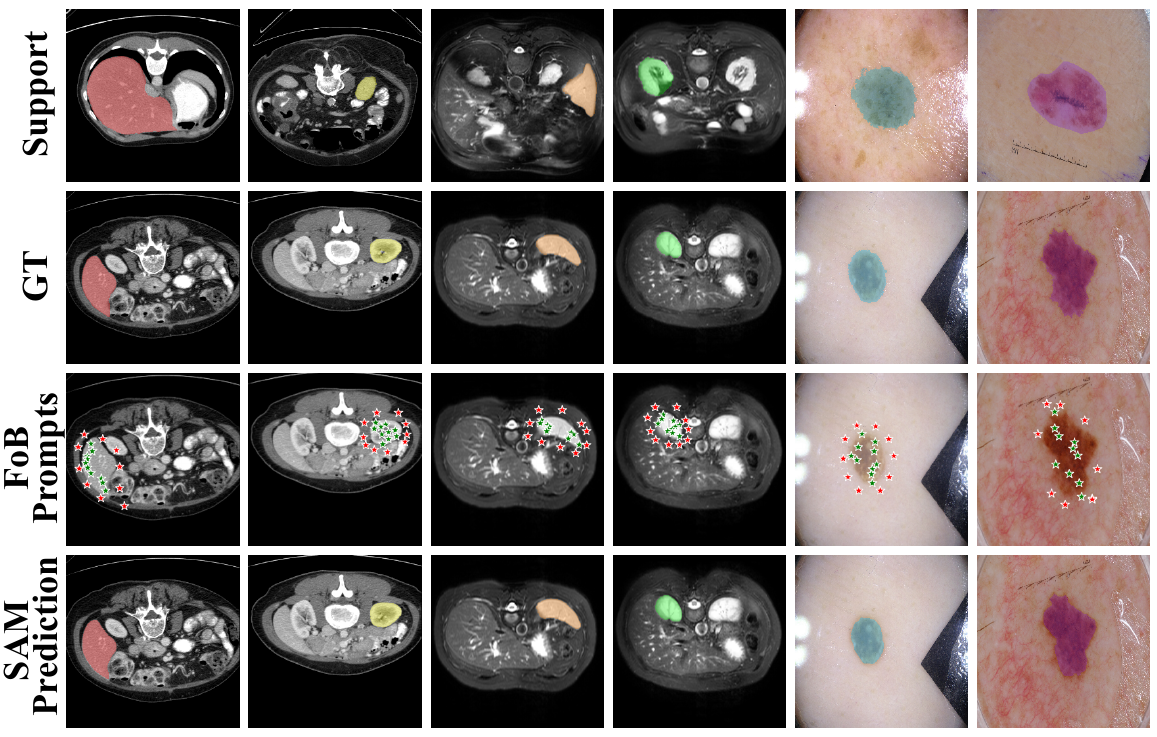}
    \vspace{-1ex}
    \caption{Qualitative results of our method across different datasets. See Appendix \ref{appendix:visualization} for more visualizations.}
    \label{fig:main_tex_segmentation}
            \vspace{-3ex}
\end{figure}

These results indicate that our method significantly improves FSMIS by incorporating SAM compared to conventional FSMIS models, and mitigates over-segmentation through background prompts compared to the previous SAM-based methods. Figure \ref{fig:main_tex_segmentation} and Appendix \ref{appendix:visualization} present qualitative results produced by our method.

\begin{table}[!t]
    \centering
    \vspace{-2ex}
        \caption{Ablation studies of model components on Dice score (\%). The best results are indicated in bold.}
        \vspace{-1ex}
\resizebox{0.9\linewidth}{!}{
    \begin{tabular}{>{\centering\arraybackslash}p{0.07\linewidth}>{\centering\arraybackslash}p{0.07\linewidth}>{\centering\arraybackslash}p{0.07\linewidth}|>{\centering\arraybackslash}p{0.08\linewidth}>{\centering\arraybackslash}p{0.08\linewidth}>{\centering\arraybackslash}p{0.08\linewidth}>{\centering\arraybackslash}p{0.08\linewidth}>{\centering\arraybackslash}p{0.08\linewidth}}
        \toprule
        BPPC& BCM& SPR& Liv & RK & LK & Spl & Avg \\ 
        \midrule
        \checkmark&  &   & 81.58& 83.21& 79.88& 79.38& 81.01\\
        \checkmark&  & \checkmark& 84.2& 86.27& 82.13& 81.35& 83.49\\
        \checkmark& \checkmark&   & 84.87& \textbf{87.61}& 84.01& 83.99& 85.12\\
        \checkmark& \checkmark& \checkmark& \textbf{86.51}& 86.51& \textbf{87.29}& \textbf{84.54}& \textbf{86.21}\\
        \bottomrule
    \end{tabular}
    }
    \label{table:module}
\end{table}

\begin{figure}[!t]
    \centering
    \includegraphics[width=0.95
    \linewidth]{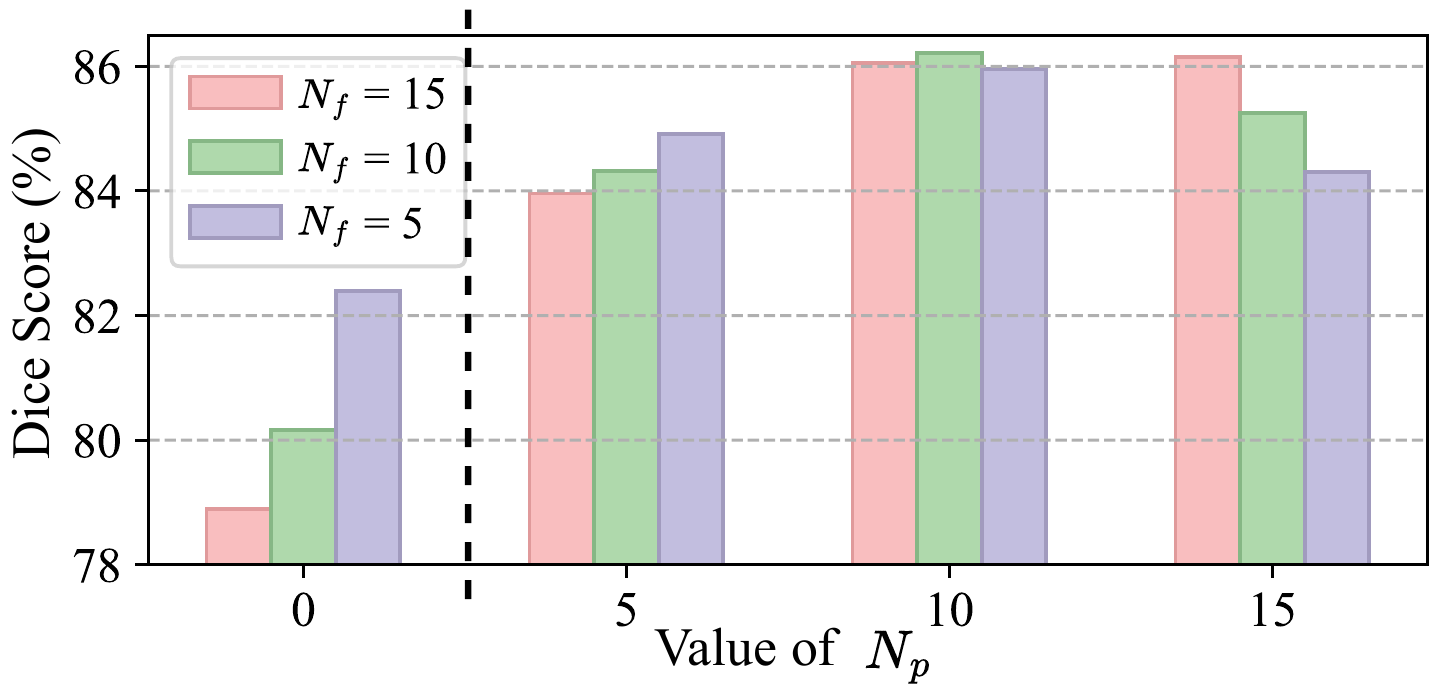}
    \vspace{-2ex}
    \caption{Analysis of the values of $N_f$ and $N_p$.}
    \label{fig:num_point}
    \vspace{-1ex}
\end{figure}

\subsection{FoB as a Domain-robust Prompt Generator}
The recently proposed Cross-domain Few-shot Medical Image Segmentation (CD-FSMIS) task \cite{ROBUSTEMD,FAMNet} aims to build models with strong generalization capability to mitigate the impact of domain shift, \eg, training on CT base categories while segmenting novel MRI categories. We evaluate our FoB on this task and compare it with SOTA methods, including RobustEMD \cite{ROBUSTEMD} and FAMNet \cite{FAMNet}. As shown in Table \ref{table:cross_domain}, FoB provides accurate prompts across domains and significantly outperforms methods tailored for CD-FSS when combined with SAM. We attribute this to FoB’s emphasis on contextual information and geometric positioning through point-level matching, both of which are domain-invariant and thus transferable across domains.


\subsection{Ablation Studies}
The ablation studies were conducted on the Abd-CT dataset. More ablations are  in Appendix \ref{appendix:Why_FoB} discussing why coarse-mask prompting is insufficient and Appendix \ref{appendix:ablation_studies}, including design choices and hyperparameter analyses.

\vspace{0.1cm}
\noindent \textbf{Effect of Model Components.}
Table \ref{table:module} shows that both BCM and SPR significantly enhance the final segmentation accuracy. Notably, BCM improves performance by 4.11\%, indicating that context reasoning effectively facilitates the localization of background prompts. Thus, SPR further boosts accuracy by 1.09\% through correction of the predicted prompts. Notably, BPPC is indispensable, as it provides the foundation for prompt localization.

\vspace{0.1cm}
\noindent \textbf{Number of Prompts.}
From Figure \ref{fig:num_point}, setting both the number of foreground and background prompts to 10 leads to the best result. Notably, regardless of the foreground prompts number ($N_f$), having background prompts (right of the dashed line) consistently improves the performance compared to the case without background prompts (left of the dashed line), indicating that accurate background prompts from FoB prevent SAM’s over-segmentation.

\vspace{0.1cm}
\noindent \textbf{Loss Functions.} 
As shown in Table~\ref{table:loss}, $\mathcal{L}_{fore}$ is essential for accurate foreground prompts. Removing $\mathcal{L}_{heat}$ and using only coordinate regression via $\mathcal{L}_{coor}$ yields poor results due to increased learning complexity \cite{XY_COMPLEXITY}. \ $\mathcal{L}_{coor}$ is indispensable as it is the only supervision for SPR. Moreover, the contrastive loss $\mathcal{L}_{rac}$ improves results by 2.28\% by preventing background prompts from falling into foreground.

\begin{figure}[t]
    \centering
    \vspace{-1.5ex}
    \includegraphics[width=0.90\linewidth]{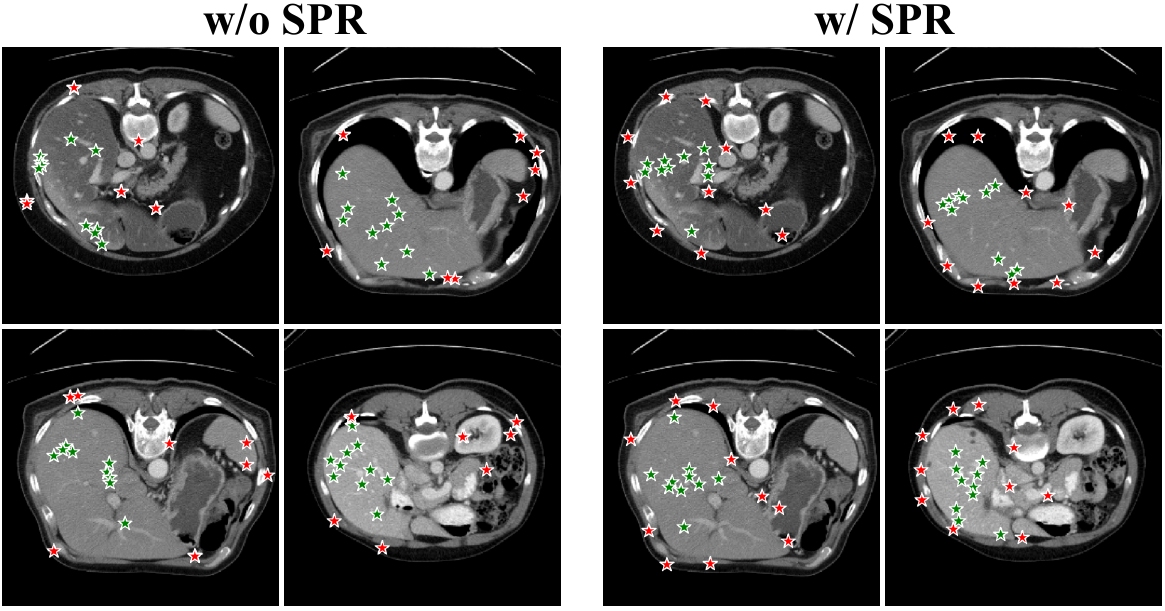}
    \vspace{-1ex}
    \caption{Visualization of the effectiveness of SPR.}
    \label{fig:main_SPR_EFFECT}
    \vspace{-2ex}
\end{figure}

\vspace{0.1cm}
\noindent \textbf{Effect of SPR.}
We visualize the effect of SPR on refining the background prompt structure by comparing the predictions with and without SPR in Figure~\ref{fig:main_SPR_EFFECT}. The model with SPR effectively learns to predict smooth, ring-like background prompts that closely follow the target shape. In contrast, several flaws are observed when SPR is removed. This demonstrates the necessity of SPR for preserving the distributional structure among background prompts.

\begin{table}[!t]
    \centering
        \vspace{-2ex}
    \caption{Ablation studies of the loss functions on Dice score (\%). The best results are indicated in bold.}
            \vspace{-1ex}
\resizebox{0.95\linewidth}{!}{
    \begin{tabular}{>{\centering\arraybackslash}p{0.05\linewidth}>{\centering\arraybackslash}p{0.05\linewidth}>{\centering\arraybackslash}p{0.05\linewidth}>{\centering\arraybackslash}p{0.06\linewidth}|>{\centering\arraybackslash}p{0.08\linewidth}>{\centering\arraybackslash}p{0.08\linewidth}>{\centering\arraybackslash}p{0.08\linewidth}>{\centering\arraybackslash}p{0.08\linewidth}>{\centering\arraybackslash}p{0.08\linewidth}}
        \toprule
        $\mathcal{L}_{heat}$& $\mathcal{L}_{coor}$& $\mathcal{L}_{fore}$& $\mathcal{L}_{rac}$& Liv & RK & LK & Spl & Avg \\ 
        \midrule
        &  \checkmark &  \checkmark &  \checkmark & 59.24& 76.92& 77.63& 68.01& 70.45\\
        \checkmark & \checkmark &  & \checkmark & 34.01& 33.92& 35.17& 37.95& 35.26\\
        \checkmark & \checkmark & \checkmark & & 83.67& 83.71& 83.02& 85.33& 83.93\\
         \checkmark & \checkmark & \checkmark & \checkmark & \textbf{86.51}& \textbf{86.51}& \textbf{87.29}& \textbf{84.54}& \textbf{86.21}\\
        \bottomrule
    \end{tabular}
    }

    \label{table:loss}
\end{table}

\section{Conclusions}
FoB is a novel background-centric prompt generator that constrains SAM’s over-segmentation and fully unleashes its potential in FSMIS. By reformulating SAM-based segmentation as a background-centric prompting problem, FoB exploits the contextual and structural priors of medical images to generate highly accurate and generalizable background prompts. Extensive experiments on diverse modalities show that FoB consistently outperforms existing FSMIS and SAM-based methods, achieving state-of-the-art performance. FoB also enjoys strong generalization in prompt generation  under cross-domain settings and FoB provides an efficient, plug-and-play solution to enhance the clinical applicability of foundation models.

\section*{Acknowledgments} This work was supported by the National Natural Science Foundation of China (NSFC) under the Grants No. 62371235 and No. U25A20444.

{
    \small
    \bibliographystyle{ieeenat_fullname}
    \bibliography{main}

\begin{thebibliography}{9}
\providecommand{\natexlab}[1]{#1}
\providecommand{\url}[1]{\texttt{#1}}
\expandafter\ifx\csname urlstyle\endcsname\relax
  \providecommand{\doi}[1]{doi: #1}\else
  \providecommand{\doi}{doi: \begingroup \urlstyle{rm}\Url}\fi

\bibitem[Achanta et~al.(2012)Achanta, Shaji, Smith, Lucchi, Fua, and
  Süsstrunk]{supp_slic}
Radhakrishna Achanta, Appu Shaji, Kevin Smith, Aurelien Lucchi, Pascal Fua, and
  Sabine Süsstrunk.
\newblock Slic superpixels compared to state-of-the-art superpixel methods.
\newblock \emph{IEEE Transactions on Pattern Analysis and Machine
  Intelligence}, 34\penalty0 (11):\penalty0 2274--2282, 2012.

\bibitem[Ayzenberg et~al.(2024)Ayzenberg, Giryes, and Greenspan]{supp_protosam}
Lev Ayzenberg, Raja Giryes, and Hayit Greenspan.
\newblock Protosam: One shot medical image segmentation with foundational
  models.
\newblock \emph{arXiv preprint arXiv:2407.07042}, 2024.

\bibitem[Cheng et~al.(2023)Cheng, Ye, Deng, Chen, Li, Wang, Su, Huang, Chen,
  Sun, He, Zhang, Zhu, and Qiao]{supp_SAMMED2D}
Junlong Cheng, Jin Ye, Zhongying Deng, Jianpin Chen, Tianbin Li, Haoyu Wang,
  Yanzhou Su, Ziyan Huang, Jilong Chen, Lei Jiangand~Hui Sun, Junjun He,
  Shaoting Zhang, Min Zhu, and Yu Qiao.
\newblock Sam-med2d.
\newblock \emph{arXiv preprint arXiv:2308.16184}, 2023.

\bibitem[Dosovitskiy et~al.(2021)Dosovitskiy, Beyer, Kolesnikov, Weissenborn,
  Zhai, Unterthiner, Dehghani, Minderer, Heigold, Gelly, Uszkoreit, and
  Houlsby]{supp_VIT}
Alexey Dosovitskiy, Lucas Beyer, Alexander Kolesnikov, Dirk Weissenborn,
  Xiaohua Zhai, Thomas Unterthiner, Mostafa Dehghani, Matthias Minderer, Georg
  Heigold, Sylvain Gelly, Jakob Uszkoreit, and Neil Houlsby.
\newblock An image is worth 16x16 words: Transformers for image recognition at
  scale.
\newblock In \emph{ICLR}, 2021.

\bibitem[Gu and Dao(2023)]{supp_mamba}
Albert Gu and Tri Dao.
\newblock Mamba: Linear-time sequence modeling with selective state spaces.
\newblock \emph{arXiv preprint arXiv:2312.00752}, 2023.

\bibitem[Kirillov et~al.(2023)Kirillov, Mintun, Ravi, Mao, Rolland, Gustafson,
  Xiao, Whitehead, Berg, Lo, et~al.]{supp_Sam}
Alexander Kirillov, Eric Mintun, Nikhila Ravi, Hanzi Mao, Chloe Rolland, Laura
  Gustafson, Tete Xiao, Spencer Whitehead, Alexander~C Berg, Wan-Yen Lo, et~al.
\newblock Segment anything.
\newblock In \emph{ICCV}, pages 4015--4026, 2023.

\bibitem[Ma et~al.(2024)Ma, He, Li, Han, You, and Wang]{supp_MedSAM}
Jun Ma, Yuting He, Feifei Li, Lin Han, Chenyu You, and Bo Wang.
\newblock Segment anything in medical images.
\newblock \emph{Nature Communications}, 15:\penalty0 654, 2024.

\bibitem[Ouyang et~al.(2020)Ouyang, Biffi, Chen, Kart, Qiu, and
  Rueckert]{supp_ssl-alp}
Cheng Ouyang, Carlo Biffi, Chen Chen, Turkay Kart, Huaqi Qiu, and Daniel
  Rueckert.
\newblock Self-supervision with superpixels: Training few-shot medical image
  segmentation without annotation.
\newblock In \emph{ECCV}, pages 762--780. Springer, 2020.

\bibitem[Ye et~al.(2023)Ye, Cheng, Chen, Deng, Li, Wang, Su, Huang, Chen,
  Jiang, Sun, Zhu, Zhang, He, and Qiao]{supp_SA-MED2D-20M}
Jin Ye, Junlong Cheng, Jianpin Chen, Zhongying Deng, Tianbin Li, Haoyu Wang,
  Yanzhou Su, Ziyan Huang, Jilong Chen, Lei Jiang, Hui Sun, Min Zhu, Shaoting
  Zhang, Junjun He, and Yu Qiao.
\newblock Sa-med2d-20m dataset: Segment anything in 2d medical imaging with 20
  million masks, 2023.

\end{thebibliography}
}


\appendix
\let\thetitle\relax
\title{Focus on Background: Exploring SAM’s Potential in Few-shot Medical Image Segmentation with Background-centric Prompting\\ -- \textit{Supplementary Material} -- }

\maketitle

\begingroup
\renewcommand\thefootnote{}
\footnotetext{ \hspace{-3ex} \Letter\ Corresponding authors.}
\endgroup


\begin{figure}[t]
    \centering
    \includegraphics[width=\linewidth]{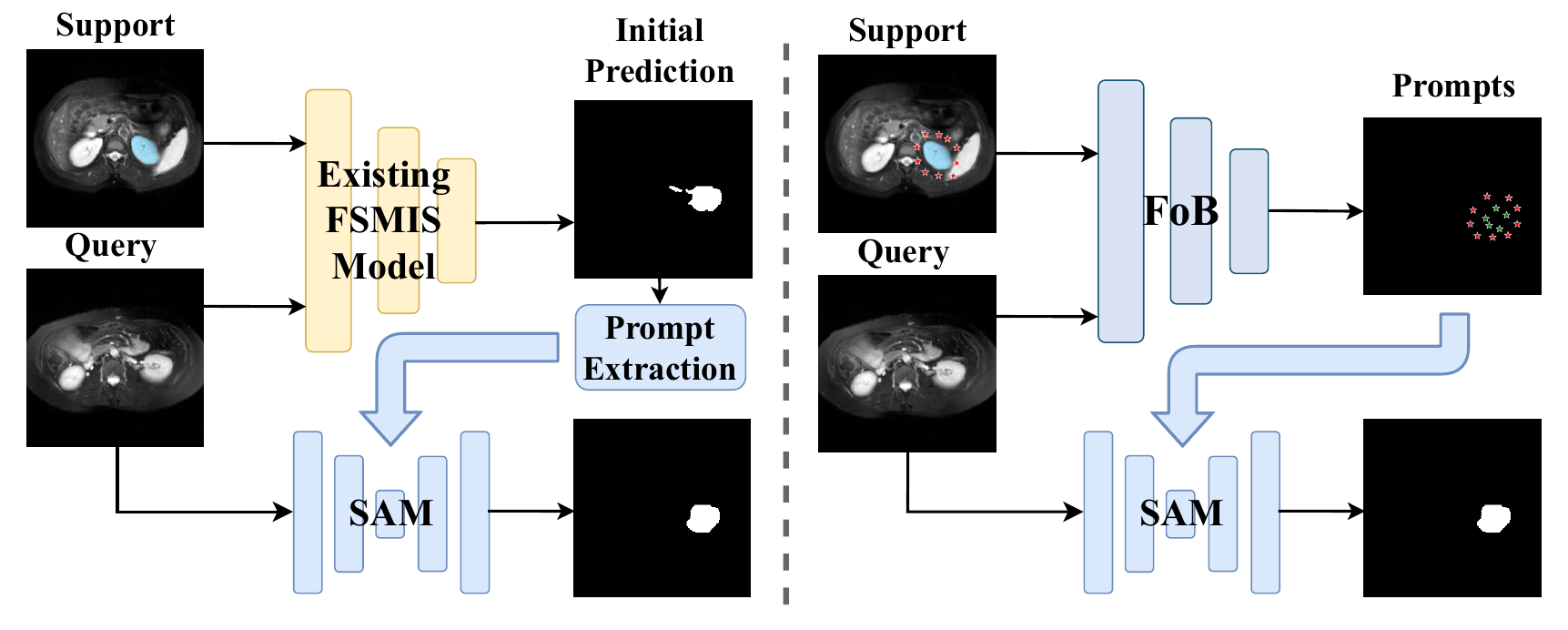}
    \caption{
    Comparison of the previous method, ProtoSAM  \protect\citelatex{supp_protosam} (left), and our method (right). ProtoSAM connects an existing FSMIS model with SAM by extracting prompts from the coarse segmentation output of the FSMIS model, which often fails to provide accurate background prompts due to inherent flaws. In contrast, our method  predicts both precise background and foreground prompts,  improving SAM’s use in medical image segmentation.
}
    \label{fig:method_compare}
    \vspace{-0.3cm}
\end{figure}

\section{Discussion: Why FoB Instead of Coarse Mask-based Prompting}
\label{appendix:Why_FoB}
Our method introduces a dedicated prompt generator specifically designed for SAM-based automatic segmentation.  Figure \ref{fig:method_compare} shows this design fundamentally differs from previous approaches such as ProtoSAM \citelatex{supp_protosam}, which simply combines an off-the-shelf FSMIS model (SSL-ALPNet \citelatex{supp_ssl-alp}) with SAM and extracts prompts directly from the coarse predictions of the FSMIS model. Extracting prompts in this manner follows two strategies (see Figure \ref{fig:why_FoB}) which we analyze  below and explain their suboptimality:

\begin{figure}[t]
    \centering
    \includegraphics[width=\linewidth]{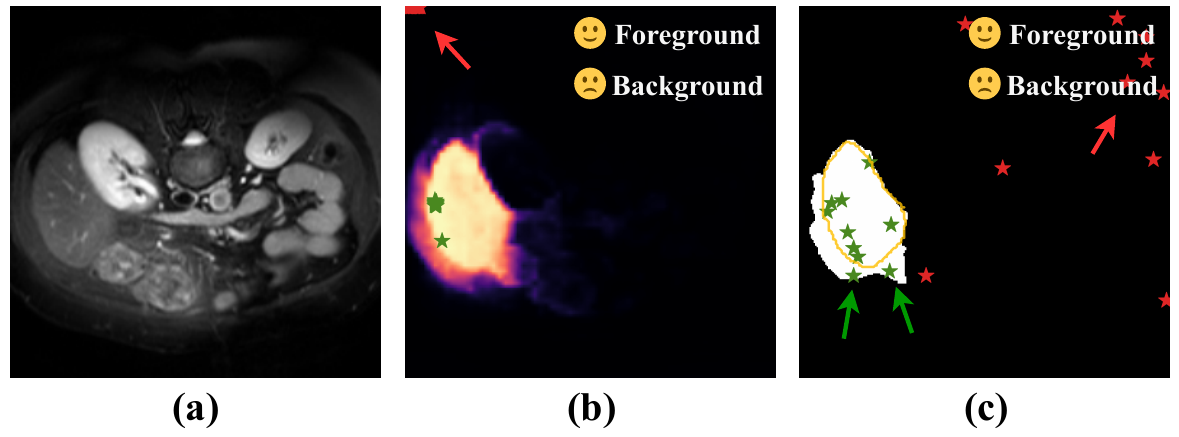}
    \caption{Previous prompt extraction methods based on existing FSMIS model outputs. 
    (a) Original image.
    (b) Prediction confidence-based method, which selects high-confidence foreground and background points as prompts.
    (c) Coarse binary mask-based method, which randomly samples prompts within the predicted foreground and background regions of the mask. The yellow line indicates the ground truth. The arrows in (b) and (c) indicate incorrect prompt locations.
}
    \label{fig:why_FoB}
    \vspace{-0.3cm}
\end{figure}

\renewcommand{\labelenumi}{\roman{enumi}.}
\begin{enumerate}[leftmargin=0.4cm]
   \item {\bf Extracting prompts based on prediction confidence~\citelatex{supp_protosam}.} Figure \ref{fig:why_FoB}(b) shows that boundaries are hard to be accurately delineated. High-confidence background prompts selected by this method are not useful as they tend to remain far away from object boundaries. This contradicts our objective, as ideal background prompts should be placed adjacent to the outer boundary, guiding SAM to more effectively discriminate foreground from background to suppress over-segmentation.
    \item {\bf Extracting prompts directly from the coarse binary mask.} Figure~\ref{fig:why_FoB}(c) shows due to the inherent limitations of pseudo-label supervision and the generalization capacity of FSMIS, the resulting coarse predictions are inaccurate. This purely geometric sampling strategy lacks semantic/shape awareness of the object. Thus,  background prompts may fail to reach the boundaries, and foreground prompts may mistakenly be placed in background regions, or vice versa.
\end{enumerate}

\noindent
In contrast, FoB is directly supervised to predict 
boundary- adjacent background prompts to achieve accurate guidance. FoB models 
relations among prompts by transformer, producing background prompts with more coherent placement and shape that  conforms well to the true object boundaries.

\section{Additional Ablation Studies}
\label{appendix:ablation_studies}

\begin{table}[t]
\centering
\resizebox{\linewidth}{!}{
\begin{tabular}{c c >{\centering\arraybackslash}p{0.1\linewidth}>{\centering\arraybackslash}p{0.15\linewidth}}
\toprule
\multirow{2}{*}{Model} & \multirow{2}{*}{mDice} & Param.& Latency \\
& & (M) & (ms/img) \\
\midrule
FoB-Transformer & 86.21 & 49.31 & 57.49 \\
FoB-Mamba       & 85.31 & 49.90 & 33.95 \\
\bottomrule
\end{tabular}
}
\caption{Comparison of Transformer- and Mamba-based implementations of the BCM module. “mDice” denotes the mean Dice score (\%), and “Param.” indicates the number of parameters.}
\label{tab:mamba}
\end{table}

\begin{table}[t]
\centering
\begin{tabular}{c|ccccc}
\toprule
$r$ & Liver & RK & LK & Spleen & Mean \\ \midrule
19 & 82.12& 83.41& 82.28& 84.01& 82.96\\
15 & \textbf{86.51}& 86.51& \textbf{87.29}& 84.54& \textbf{86.21}\\
11 & 86.37& \textbf{87.33}& 86.12& \textbf{84.93}& 86.19\\
7 & 79.21& 82.66& 84.76& 80.47& 81.78\\
\bottomrule
\end{tabular}
\caption{Ablation study (Dice score \% reported) on the impact of the dilation kernel size $r$. Smaller $r$ moves background prompts closer to the foreground.}
\label{tab:dilation_r}
 \vspace{-0.3cm}
\end{table}

\subsection{Design Choices}
\subsubsection{Transformer \vs Mamba for Background-centric Context Modeling.}
The recently proposed architecture, Mamba \citelatex{supp_mamba}, is a competitive alternative to Transformers. In BCM, we treat pixel features as tokens and apply a Transformer for  context modeling based on self-attention.  Mamba models sequence dependencies through a learnable state-space transition. To investigate Mamba's  contextual reasoning, we re-implement FoB by replacing the multi-head self-attention in BCM with Mamba. The resulting performance-efficiency trade-off is summarized in Table \ref{tab:mamba}. We observe that both implementations achieve comparable segmentation performance and exhibit similar parameter scales, indicating their equivalent potential for clinical deployment. However, the Mamba-based BCM demonstrates significantly lower latency due to its linear-time inference complexity. This advantage enables faster segmentation feedback on resource-constrained medical devices. 

\subsubsection{Discussion: Are Medical SAMs Suitable for This Task?}
As shown in Table \ref{table:main} of the main text, SAM-Med2D \citelatex{supp_SAMMED2D}  underperforms in comparison to the vanilla SAM on abdominal datasets, while significantly outperforming it on Skin-DS. Table \ref{table:medsam}  presents the results of using FoB to prompt another popular SOTA medical SAM, MedSAM \citelatex{supp_MedSAM}, which was not evaluated in the main text due to its support for box prompts only. The results  show suboptimal performance which we attribute  to the architectural choice: both SAM-Med2D and MedSAM adopt the ViT-B (base) backbone, which is less expressive than the ViT-H (huge) backbone used in the original SAM \citelatex{supp_Sam}. Notably, most SOTA medical SAM variants, including SAM-Med2D and MedSAM, rely on ViT-B, due to its reduced data requirements \citelatex{supp_VIT}, which is a property well aligned with the limited availability of annotated medical data.

\begin{figure}[t]
    \centering
    \includegraphics[width=\linewidth]{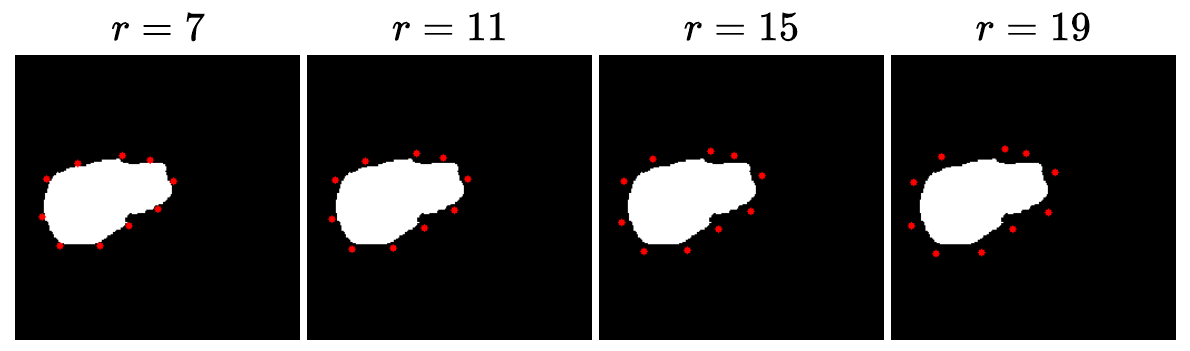}
    \caption{Visualization of support background prompts generated with different $r$ in BPPC. Best viewed under large zoom.}
    \label{fig:dilation_proximity}
    \vspace{-0.3cm}
\end{figure}

The superior performance of SAM-Med2D on some datasets may stem from data overlap between training and testing. Given the scarcity of public medical datasets, it is possible that some test images, especially from Skin-DS, were seen during the training of SA-Med2D-20M \citelatex{supp_SA-MED2D-20M}, which was used to train SAM-Med2D. 

In contrast, the few-shot segmentation setting strictly prohibits access to target classes during training. Our intention is to leverage the category-agnostic segmentation ability of the SAM trained on natural images and extend its generalization to the medical domain, whereas using medical SAMs risks violating the fundamental assumptions of few-shot setting. Thus, while we report results from SAM-Med2D and MedSAM for completeness and to assess prompt generalization across model variants, we advocate using the vanilla SAM to comply with the FSMIS protocol.

\begin{table*}[t]
\centering
\vspace{-1ex}
\renewcommand\arraystretch{1.2}
\resizebox{\textwidth}{!}{
\begin{tabular}{c | c c c c c | c c c c c | c c c c } 
\toprule
   \multirow{2}{*}{Setting}& \multicolumn{5}{c|}{Abd-CT} & \multicolumn{5}{c|}{Abd-MRI} & \multicolumn{4}{c}{Skin-DS} \\
  & Liv& RK & LK & Spl& Avg. & Liv& RK & LK & Spl& Avg. & Mel& Nev& SK & Avg. \\
\midrule
 \textbf{Setting I}& 69.51& 65.79& 73.01& 57.93& 66.56& 71.54& 78.57& 80.46& 68.81& 74.85& 71.59& 75.34& 68.75& 71.89\\
 \midrule
 \textbf{Setting II}& 68.61& 64.30& 67.67& 58.68& 64.82& 67.99& 75.57& 74.47& 62.86& 70.22& 72.91& 71.25& 68.49& 70.88\\
\bottomrule
\end{tabular}
}
\caption{Segmentation results of MedSAM when prompted by FoB. Bounding box prompts are obtained by computing the minimum enclosing rectangle of the background prompts generated by FoB. These results further validate that medical SAMs are in fact unsuitable for FSMIS tasks, due to inherent architectural limitations and potential violations of the few-shot learning protocol.}
\label{table:medsam}
\vspace{-3ex}
\end{table*}

\subsection{Hyperparameter Settings}
\subsubsection{Optimal Proximity of Background Prompts.}
We design FoB to predict background prompts that are located close to the foreground region, thereby constraining the over-segmentation errors that extend beyond object boundaries. This section investigates an important question: ``Is closer always better?''

In our design, this distance is controlled by a dilation kernel size $r$. A smaller $r$ generates background prompts closer to the foreground in BPPC. Figure \ref{fig:dilation_proximity} shows support background prompts with different $r$. 
As shown in Table \ref{tab:dilation_r}, reducing the distance between the background prompts and the foreground gradually improves the segmentation accuracy of SAM, indicating that over-segmentation is effectively suppressed. However, when this distance becomes excessively small, the performance drops. We attribute this to the model excessively prioritizing proximity, which increases the risk of background prompts falling inside the true foreground,  introducing conflicting signals and deteriorating the segmentation quality.

\begin{table}[t]
\centering
\begin{tabular}{c|ccccc}
\toprule
$\sigma$ & Liver & RK & LK & Spleen & Mean \\ \midrule
8 & 80.02& 84.94& 85.78& 80.21& 82.74\\
4 & \textbf{86.51}& \textbf{86.51}& \textbf{87.29}& \textbf{84.54}& \textbf{86.21}\\
2 & 79.76& 75.56& 75.97& 79.91& 77.80\\
1 & 76.33& 61.93& 64.46& 62.57& 66.32\\
\bottomrule
\end{tabular}
\caption{Ablation study (Dice score \% used) on the impact of the standard deviation $\sigma$ for generating heatmaps.}
\label{table:sigma}
\end{table}

\begin{table}[t]
\centering
\begin{tabular}{c|ccccc}
\toprule
$k$ & Liver & RK & LK & Spleen & Mean \\ \midrule
16& 84.96& 76.80& 88.33& 83.89& 83.50\\
8& 86.51& \textbf{86.51}& 87.29& 84.54& \textbf{86.21}\\
4& \textbf{87.50}& 84.64& 85.58& \textbf{86.19}& 85.98\\
2& 81.05& 85.40& \textbf{88.63}& 83.62& 84.68\\
\bottomrule
\end{tabular}
\caption{Ablation study (in Dice score \%) on the impact of the deformable receptive field size $k$ in SPR.}
\label{table:k}
\vspace{-0.3cm}
\end{table}

\subsubsection{Additional Hyperparameter Analysis.}
We conduct extensive ablation studies on key hyperparameters in our model, including the standard deviation $\sigma$ for heatmap generation, the number of deformable iterations $\kappa$ and deformable receptive field size $k$ in SPR, the foreground sampling threshold $\mathcal{T}$, and the temperature parameter $\tau$ in the RAC loss. Among these, $\sigma$ has the most significant impact on model performance, as shown in Table \ref{table:sigma}. It primarily affects both the supervision signal and the prompt prototype generation. A large $\sigma$ weakens the supervision strength and produces prototypes that fail to accurately represent the prompts. Conversely, a small $\sigma$ makes the model harder to optimize and results in prototypes that are highly sensitive to noise.
Tables \ref{table:k} and \ref{table:h} present the hyperparameter analysis of the SPR module. The results show minor performance variance \wrt $\kappa$ and $k$. 
The effects of $\mathcal{T}$ and $\tau$ are summarized in Tables \ref{table:T} and \ref{table:tau}, respectively. Notably, a smaller $\tau$ corresponds to stronger contrastive constraints. As $\tau$ decreases, the segmentation performance of SAM steadily improves, suggesting that stronger feature discrimination can effectively prevent background prompts from being misclassified as foreground.
Overall, the optimal hyperparameter configuration in our experiments is $\sigma = 4$, $\kappa = 3$, $k = 8$, $\mathcal{T} = 0.90$, and $\tau = 0.10$.

\begin{table}[t]
\centering
\begin{tabular}{c|ccccc}
\toprule
$\kappa$& Liver & RK & LK & Spleen & Mean \\ \midrule
7& 82.62& 83.99& 83.94& 77.67& 82.06\\
5& 84.43& \textbf{86.68}& 86.63& 83.75& 85.37\\
3& \textbf{86.51}& 86.51& 87.29& 84.54& \textbf{86.21}\\
1& 84.77& 84.17& \textbf{87.42}& \textbf{84.68}& 85.26\\
\bottomrule
\end{tabular}
\caption{Ablation study (Dice score \% used) on the impact of the number of deformable iterations $\kappa$ in SPR.}
\label{table:h}
\end{table}

\begin{table}[t]
\centering
\begin{tabular}{c|ccccc}
\toprule
$\mathcal{T}$ & Liver & RK & LK & Spleen & Mean \\ \midrule
0.95 & \textbf{88.75}& 86.03& 83.99& \textbf{85.27}& 86.01\\
0.90 & 86.51& \textbf{86.51}& \textbf{87.29}& 84.54& \textbf{86.21}\\
0.85 & 87.94& 83.77& 85.62& 82.47& 84.95\\
0.80 & 87.38& 84.55& 84.01& 81.79& 84.43\\
\bottomrule
\end{tabular}
\caption{Ablation study (Dice score \% used) on the impact of the foreground prompt sampling threshold $\mathcal{T}$.}
\label{table:T}
\end{table}

\begin{table}[t]
\centering
\begin{tabular}{c|ccccc}
\toprule
$\tau$ & Liver & RK & LK & Spleen & Mean \\ \midrule
0.7 & 81.74& 86.31& 85.72& 80.09& 83.47\\
0.5 & 84.96& \textbf{86.79}& 84.94& 82.01& 84.68\\
0.3 & \textbf{87.24}& 86.37& 86.97& 83.17& 85.94\\
0.1 & 86.51& 86.51& \textbf{87.29}& \textbf{84.54}& \textbf{86.21}\\
\bottomrule
\end{tabular}
\caption{Ablation study (Dice score \% used) on the impact of the temperature $\tau$ in the RAC loss.}
\label{table:tau}
\vspace{-0.3cm}
\end{table}

\subsection{Robustness Analysis}
\subsubsection{Does BCM rely on accurate foreground prediction?}
In BCM, we first predict a foreground mask to help subsequent modeling differentiate foreground and background regions. While accurate foreground prediction is beneficial, it is not strictly required, \ie, coarse predictions are already sufficient to localize the foreground region because BCM does not rely solely on foreground signals. It explicitly models foreground–background relationships while also leveraging additional contextual cues, \eg, anatomical structures encoded in the query spatial layout. Thus, the performance of BCM is not determined by foreground prediction alone.

\begin{figure}[t]
    \centering
    \includegraphics[width=0.99\linewidth]{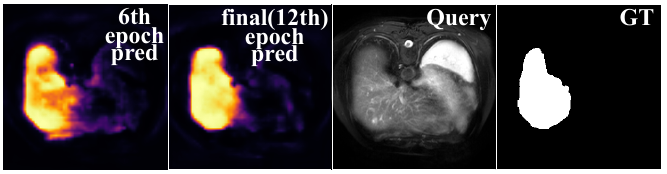}
    \caption{Evolution of the foreground probability map over training epochs. As training proceeds, the foreground prediction becomes highly accurate even under ambiguous boundaries.}
    \label{fig:fg_accuracy}
\end{figure}

Furthermore, the predicted foreground remains reliable even in scenarios where foreground and background are visually ambiguous. This is because the prediction process is supervised by $\mathcal{L}_{fore}$, which encourages semantically discriminative feature learning and promotes precise boundary delineation. In addition, $\mathcal{L}_{rac}$ explicitly focuses on hard boundary regions, further enhancing feature discriminability for foreground–background separation. 
As training progresses, the foreground prediction becomes increasingly accurate, as illustrated in Figure~\ref{fig:fg_accuracy}.

\subsubsection{Is matching robust under significant background variation?}
\begin{figure}[t]
    \centering
    \includegraphics[width=0.99\linewidth]{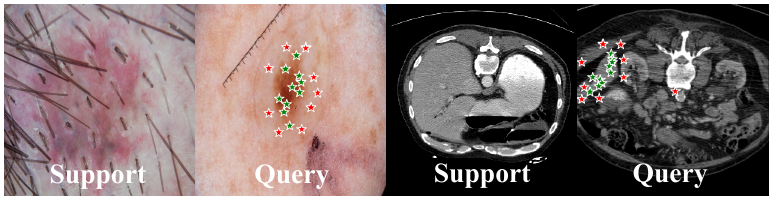}
    \caption{Visualization of matching robustness of the BCM module under significant background variations.}
    \label{fig:large_variance}
\end{figure}

In medical images, the background typically exhibits substantial variation across samples, which can hinder reliable matching for localizing background prompts. However, BCM mitigates this issue by leveraging global contextual information. In particular, it utilizes the foreground region, which is easier to predict than background prompts, as well as additional anatomical context from the query instance to perform instance-adaptive reasoning. This design makes the matching process  robust to background variations. When support prompt prototypes are less reliable due to large background discrepancies, BCM compensates by leveraging query contextual cues to refine the matching. We further demonstrate this robustness using the FoB model without SPR in Figure \ref{fig:large_variance}. BCM accurately localizes the background prompts through matching, even under significant variations.

\begin{table}[t]
\centering
\resizebox{\linewidth}{!}{
\begin{tabular}{l|ccccc}
\toprule
Graph& Liver & RK & LK & Spleen & Mean \\ \midrule
No $\mathbf{A}$& 81.32& 87.81& 82.43& 78.55& 82.53\\
$\mathbf{A}^{ring}$ only& 84.99& \textbf{88.26}& 84.09& 78.95& 84.07\\
$\mathbf{A}^{ada}$ only& 83.87& 87.00& \textbf{84.88}& 77.39& 83.29\\
$\mathbf{A}^{ada} + \mathbf{A}^{ring}$& \textbf{85.61}& 88.18& 84.76& \textbf{79.31}& \textbf{84.46}\\
\bottomrule
\end{tabular}
}
\caption{Ablation study (Dice score \% used) on different graph construction strategies.}
\label{tab:graph_ablation}
\vspace{-0.3cm}
\end{table}
\subsection{Ablations on Structural Graph in SPR}
In SPR, we construct graphs to encode the structural relationships among support background prompt prototypes, which are used to regularize the distribution of query prompt prototypes in the feature space. Specifically, we construct both an adaptive graph $\mathbf{A}^{ada}$ and a ring-prior graph $\mathbf{A}^{ring}$ for subsequent modeling.

We ablate different graph construction strategies in Table~\ref{tab:graph_ablation}. The quantitative results show that $\mathbf{A}^{ring}$ contributes the most significant performance gain, as it imposes a strong ring-shaped topological prior on the query prototypes. Combining both graphs yields the best performance, as it captures both category-specific structural relationships and a general ring-like prior.

\begin{figure}[t]
    \centering
    \includegraphics[width=\linewidth]{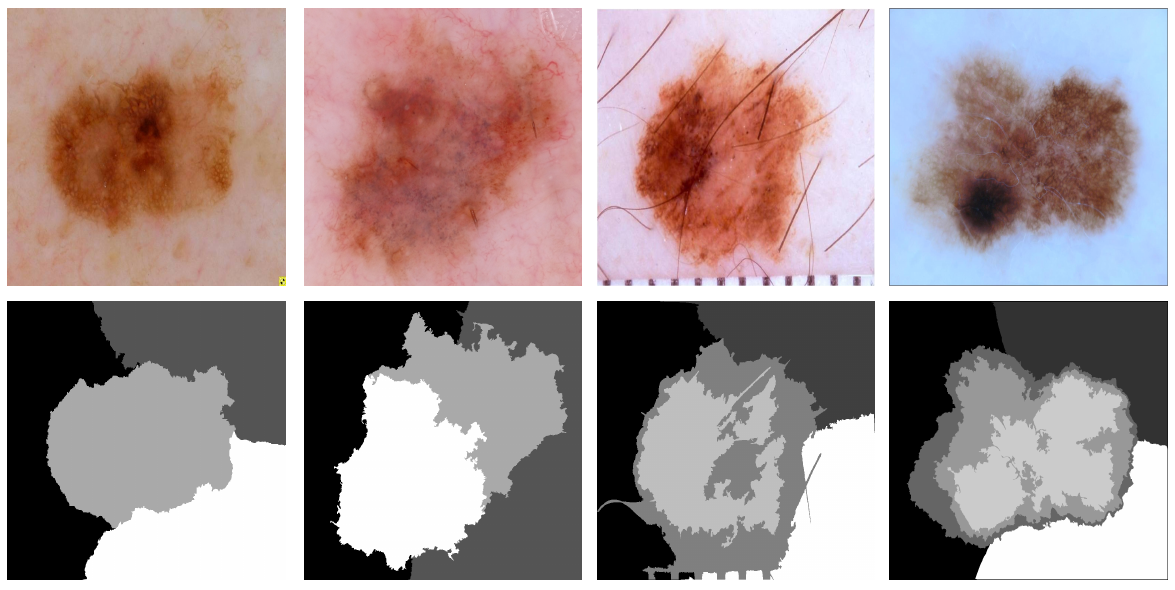}
    \caption{Illustrative examples of superpixel-based pseudo labels on Skin-DS.}
    \label{fig:ISIC_Pseudo_label}
\end{figure}

\subsection{Performance Curve}
\begin{figure}[t]
    \centering
    \includegraphics[width=0.99\linewidth]{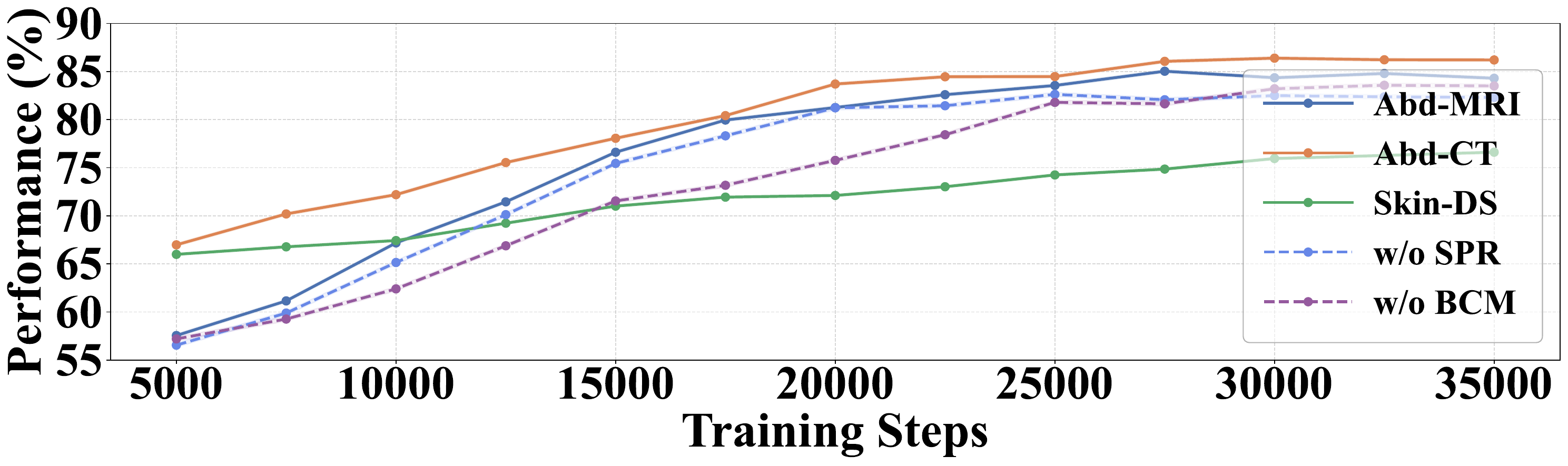}
    \caption{Performance curves of FoB \& its variants on 3 datasets.}
    \label{fig:validation_curve}
    \vspace{-0.3cm}
\end{figure}

We report the performance curves of FoB on three datasets, together with ablation result curves without BCM and SPR, as shown in Figure~\ref{fig:validation_curve}. 
We observe that on complex multi-object datasets (\eg, abdominal datasets), the initial performance is relatively low, as SAM struggles to accurately separate multiple regions under challenging scenes. 
As the background prompts become progressively more accurate, the performance improves substantially. 
The performance of our model increases steadily with the learning of background prompts. 
Furthermore, BCM accelerates convergence, while SPR consistently improves the upper performance bound, demonstrating the effectiveness of our proposed modules.

\section{Details of Superpixel-based Pseudo-labeling for Skin-DS}
\label{sup:superpix}
To the best of our knowledge, we are the first to adopt the Skin-DS dataset for FSMIS. Following the conventional pseudo-label training paradigm \citelatex{supp_ssl-alp}, we generate pseudo labels for Skin-DS to enable training FoB without requiring ground-truth annotations of skin disease in Setting I. This lets the model learn robust and generalizable patch-level features, mitigating the risk of overfitting to specific semantics and thus enhancing its ability to generalize to unseen categories during inference.
For simplicity and computational efficiency, we adopt the SLIC \citelatex{supp_slic} algorithm for pre-processing Skin-DS. SLIC performs k-means clustering in a joint color–spatial domain, yielding compact and edge-aware superpixel regions. In our implementation, the number of desired superpixels is set to 5, and the compactness parameter is set to 15. We show several processed examples in Figure \ref{fig:ISIC_Pseudo_label}. Moreover, Figure \ref{fig:visualization_skin} provides qualitative segmentation results on Skin-DS.

\section{Additional Visualizations}
\label{appendix:visualization}
\subsection{Visual Analysis on Structure-guided Prompt Refinement (Detailed)}
\begin{figure*}[t]
    \centering
    \includegraphics[width=\linewidth]{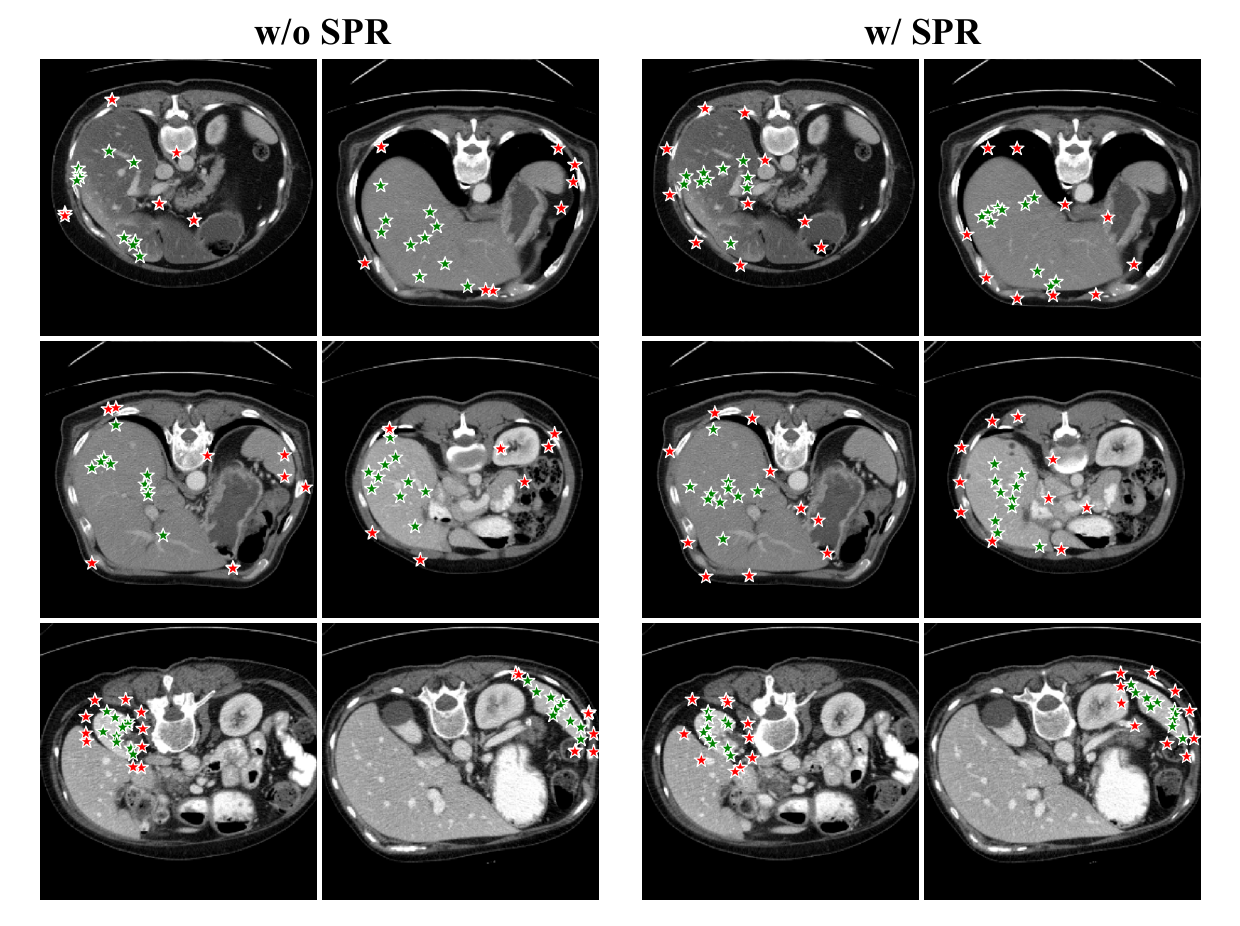}
    \caption{Qualitative effect of SPR on Abd-CT. Our proposed FoB with structure-aware refinement (w/ SPR) significantly outperforms the counterpart that solely uses BCM-predicted prompt sets (w/o SPR).}
    \label{fig:effect_of_SPR}
\end{figure*}

In Table \ref{table:module} of the main text, we quantitatively demonstrate the effectiveness of the SPR module. We further provide visualization results to highlight the significant improvements that SPR brings in generating background prompts that better align with the inherent structure. As shown in Figure \ref{fig:effect_of_SPR}, for several examples, FoB with SPR (w/ SPR) effectively learns to predict smooth, ring-like prompt distributions that closely follow the spatial shape of the target category (as indicated by support prompts), wrapping around the foreground to offer strong constraints to prevent SAM’s over-segmentation. In contrast, the predictions of the model trained without SPR are inaccurate, resulting in either outlier prompts located far from the foreground (\eg, top row, second column) or overly compact prompt clusters (\eg, bottom row, second column).

\subsection{Generated Background Prompts}
\begin{figure*}[t]
    \centering
    \includegraphics[width=\linewidth]{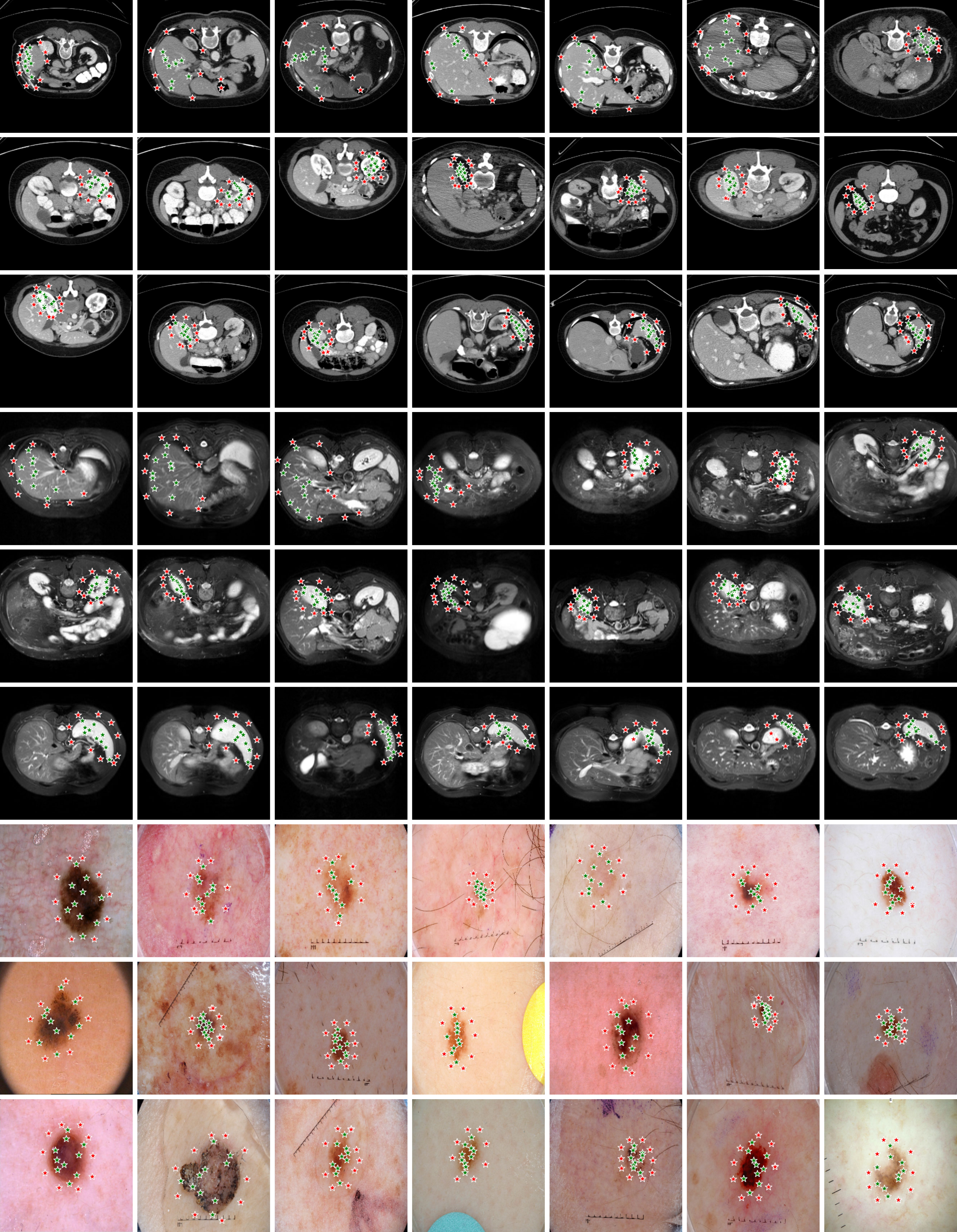}
    \caption{Visualization of prompts generated by the proposed FoB.
    Rows 1–3 correspond to Abd-CT, rows 4–6 to Abd-MRI, and rows 7–9 to Skin-DS.
    FoB produces highly reliable background prompts that play a crucial role in constraining SAM's over-segmentation.}
    \label{fig:prompt_visualization}
\end{figure*}

A comprehensive visualization of the generated background prompts by FoB across different imaging modalities is presented in Figure \ref{fig:prompt_visualization}. We observe that FoB demonstrates remarkable accuracy in localizing background points adjacent to target boundaries. These points are distributed in a morphologically consistent manner around category boundaries, offering strong guidance to constrain the over-segmentation of SAM. Moreover, FoB also yields highly accurate foreground prompts, despite relying solely on basic prototype matching without introducing any additional architectural components.

\begin{figure*}[t]
    \centering
    \includegraphics[width=\linewidth]{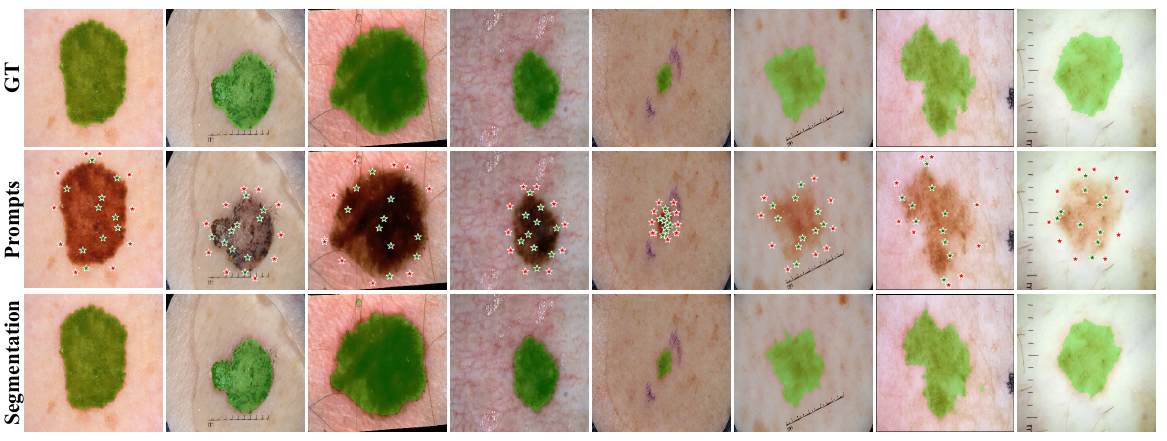}
    \caption{Qualitative segmentation results of our method on Skin-DS.}
    \label{fig:visualization_skin}
\end{figure*}


\begin{algorithm}[t]
\caption{Focus on Background Prompt Generator (1-shot).}
\label{alg:fob}
\begin{algorithmic}[1]
    \REQUIRE Support set $\mathcal{S} = \big\{(\mathbf I^{s},\mathbf M^{s})\big\}$, query image $ \mathbf I^{q}$, numbers of prompts $N_p,N_f$.
  \ENSURE Background prompts $\mathcal P_b'$, foreground prompts $\mathcal P_f$

  \STATE \textbf{Feature extraction:}
        Extract support and query features $\mathbf F^{s}$ and $ \mathbf F^{q}$ using a shared-weight encoder $f(\cdot)$.

  \STATE \textbf{Background Prompt Prototypes Construction (BPPC):}
        \begin{ALC@g}
          \STATE Sample support background prompt set $\mathcal P$ (Eq. \eqref{eq:sample_prompts}).

          \STATE Generate Gaussian heatmaps set $\mathbf{G} \!=\! [\mathbf{G}^1, \dots, \mathbf{G}^{N_p}]$ centered at each $\boldsymbol{\mu}^i \in \mathcal{P}$ (Eq. \eqref{heatmap}).
          \STATE Create background prompt prototype set $\mathbf{P}$ (Eq. \eqref{MAP}).

        \end{ALC@g}

  \STATE \textbf{Background-centric Context Modeling (BCM):}
  \begin{ALC@g}
    \STATE Get foreground suppressed query image feature $\mathbf{F}_{sup}$ (Eq. \eqref{eq4}).
    \STATE Generate coarse background prompt proposal $\boldsymbol\Phi$ with $\mathbf P$ and $\mathbf{F}_{sup}$ (Eq. \eqref{eq5}).
    \STATE Obtain the contextual modulated feature $\mathbf{F}_m$ using the masked transformer (Eq. \eqref{eq6} \& \eqref{eq7}):
    \STATE Heatmap prediction: $\hat{\mathbf{H}} \leftarrow \text{Head}(\mathbf{F}_m)$.
    \STATE Obtain coarse prompts $\mathcal{P}_b$ by selecting the maximum response in each heatmap: $\mathcal{P}_b \leftarrow \{\arg\max \hat{\mathbf{H}}^i\}_{i=1}^{N_p}$.
  \end{ALC@g}

  \STATE \textbf{Structure-guided Prompt Refinement (SPR):}
  \begin{ALC@g}
    \STATE Estimate adaptive graph $\mathbf{A}^{ada}$ with support features $\mathbf{P}$ (Eq. \eqref{eq8}).
    \STATE Compute ring prior graph $\mathbf{A}^{ring}$ (Eq. \eqref{eq9}).
    \STATE Compute $\mathbf{A}$ as a weighted sum of $\mathbf{A}^{ada}$ and $\mathbf{A}^{ring}$ (Eq. \eqref{eq10}).
    \STATE Transfer support structure to query to get $\mathbf{Q}'$ (Eq. \eqref{eq11}).
    \STATE Iteratively update prompt coordinates:
    \FOR{$i \leftarrow 1$ \TO $N_p$}
      \STATE Initialize $\mathbf{f} \leftarrow \mathbf{q}_b^{i\prime} \in \mathbf{Q}'$
      \FOR{$t \leftarrow 1$ \TO $\kappa$}
        \STATE Predict offset set  $\Delta\boldsymbol{\mu}$ (Eq. \eqref{eq12}).
        \STATE Compute weights  $\mathbf{w}$ using $\mathbf{q}_b^{i}$ (Eq. \eqref{eq13}).
        \STATE Refine location $\boldsymbol{\mu}_b^i \in \mathcal{P}_b'$ and feature $\mathbf{f}$ using $\mathbf{w}$ and $\Delta\boldsymbol{\mu}$ (Eq. \eqref{eq14} \& \eqref{eq15}).
      \ENDFOR
    \ENDFOR
  \end{ALC@g}
  \RETURN $\mathcal{P}_b'$, $\mathcal P_{f}$
\end{algorithmic}
\end{algorithm}

\subsection{Visualization of Segmentation Results}
\begin{figure*}[t]
    \centering
    \includegraphics[width=\linewidth]{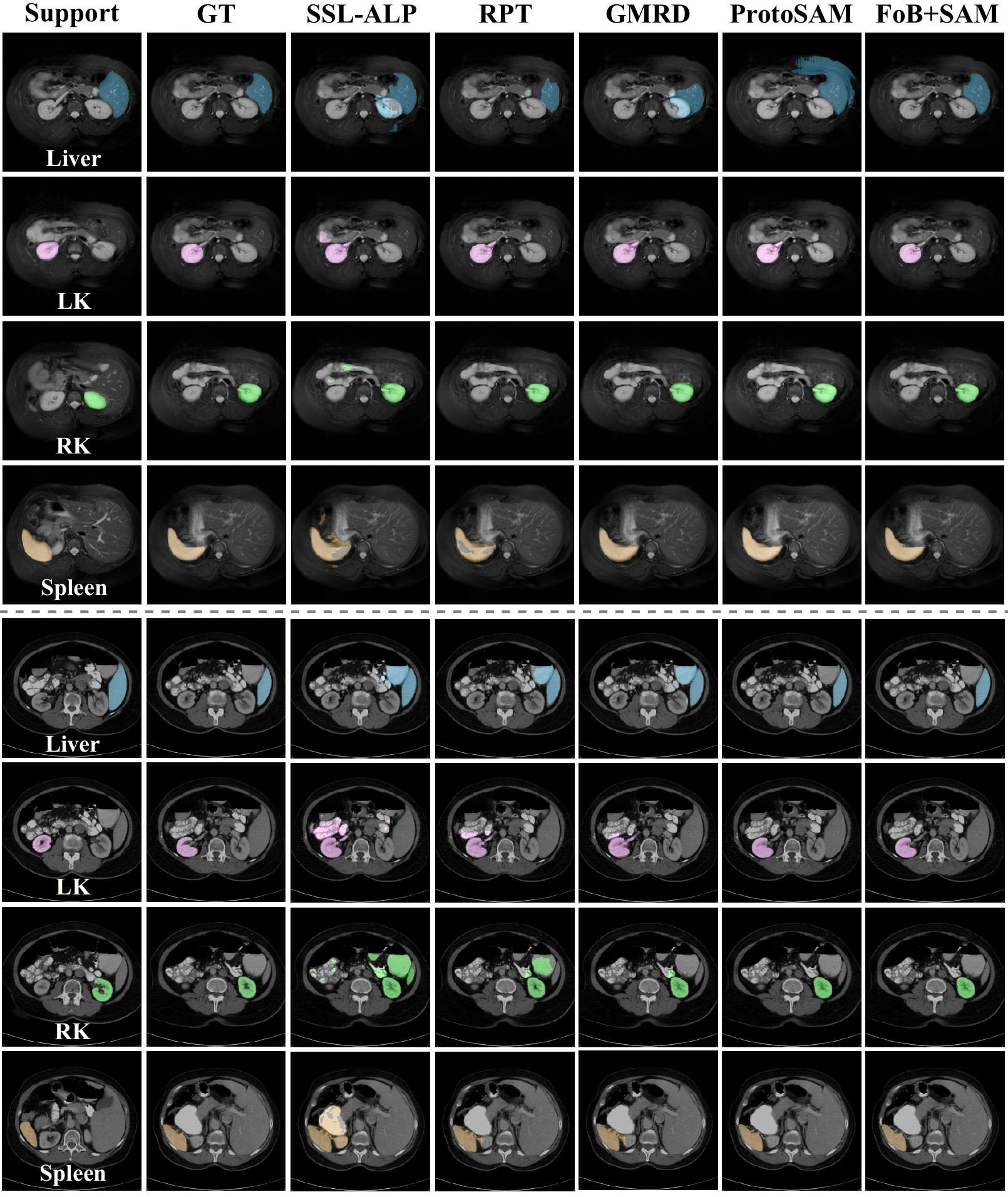}
    \caption{Qualitative comparison of segmentation results on Abd-MRI (upper) and Abd-CT (lower).}
    \label{fig:visualization_abd}
\end{figure*}

We present the qualitative results of our method in Figures \ref{fig:visualization_abd} and \ref{fig:visualization_skin}. Compared to conventional approaches and the prior SAM-based method, ProtoSAM, our approach produces more complete foreground segmentation with sharper and more decisive boundaries, benefiting from SAM's strong capability in image segmentation. It also significantly suppresses over-segmentation, a severe issue not only in ProtoSAM but also in conventional methods based on prototypical matching. Our results demonstrate the potential of background-centric few-shot SAM prompting in clinical applications, which achieves strong performance while requiring minimal annotated data.

\section{Limitations and Future Work}
Although our FoB leverages SAM to achieve accurate segmentation for common medical targets, the current design does not yet support highly irregular and thin structures, such as vessels. This limitation arises because such cases require a larger number of background prompts to avoid erroneous segmentation, whereas our design adopts a fixed number $N_p$ of background prompts. Moreover, such cases are also not well aligned with the ring-shape prior and may benefit from other advanced priors. Future work may explore supporting a dynamic number of background prompts to better adapt to the geometry and scale of target structures, as well as more adaptive strategies to correct the predicted prompts. We hope that our sparse point-matching-based paradigm can foster more SAM-based FSMIS methods and facilitate their practical deployment. 

\section{Algorithm}
Algorithm \ref{alg:fob} illustrates the proposed FoB model which comprises three key stages:
1) background prompt prototype generation from the support set via BPPC;
2) contextual modeling for enhanced background prompt localization via BCM; and
3) structure-guided refinement for calibrating erroneous query prompts via SPR.

%

{
    \small
    \bibliographystylelatex{ieeenat_fullname}
		\bibliographylatex{latex.bbl}
}




\end{document}